%% file: iclr2027_conference.tex
\documentclass{article} 
\usepackage{iclr2027_conference,times}

\input{math_commands.tex}

\usepackage{hyperref}
\usepackage{url}
\usepackage{graphicx}
\usepackage{float}
\usepackage{booktabs}
\usepackage{multirow}
\usepackage{adjustbox}
\usepackage{array}
\usepackage{comment}

\title{AnyStep-WAM: Budget-Aligned Distillation and Adaptive Inference for World Action Models}

\author{%
\makebox[\dimexpr\textwidth-2\tabcolsep\relax][c]{%
\normalfont
\begin{tabular}[t]{@{}c@{}}
\rule{0pt}{1.0em}%
\textbf{Rui Wang}$^{1,3}$\thanks{Equal contribution.}
\quad
\textbf{Xiangyu Wang}$^{1}$\footnotemark[1]
\quad
\textbf{Donglin Yang}$^{2}$
\quad
\textbf{Yibo Li}$^{1}$
\\[0.3em]
\textbf{Canyang Chen}$^{1}$
\quad
\textbf{Zhongrui Wang}$^{1}$\thanks{
Corresponding authors:
Xiaojuan Qi (\texttt{xjqi@eee.hku.hk}) and
Zhongrui Wang (\texttt{wangzr@sustech.edu.cn}).
}
\quad
\textbf{Xiaojuan Qi}$^{2,3}$\footnotemark[2]
\\[0.6em]
$^{1}$Southern University of Science and Technology
\quad
$^{2}$The University of Hong Kong
\\
$^{3}$Shenzhen Loop Area Institute
\end{tabular}%
}%
}

\iclrfinalcopy
\begin{document}

\maketitle
\lhead{}

\vspace{-25pt}
\begin{center}
\href{https://ruiwang724.github.io/AnyStep-WAM/}{Project Website}
\end{center}

\begin{abstract}
World-action models (WAMs) couple predictive visual modeling with action generation, typically relying on iterative denoising with a fixed denoising steps. 
However, manipulation tasks contain actions chunks with varying sensitivity to generation errors: critical actions require precision, while less sensitive actions allow faster generation with fewer denoising steps.
Here we introduce \textbf{AnyStep World Action Model}, a general framework for tunable-budget prediction and scene-dependent computation allocation. Our budget-aligned teacher-trajectory distillation trains interval-conditioned flow maps using explicit frozen-teacher transitions and shared low-rank adapters, supporting action generation from one-step prediction to multi-step refinement. Building on this capability, a lightweight risk-benefit scheduler predicts teacher-curvature-based difficulty and budget-specific student-teacher fidelity from a single one-step preview, selecting the smallest budget predicted to satisfy risk-adaptive fidelity requirements. We evaluate our framework on three widely used WAMs Motus, FastWAM, and LingBotVA using RoboTwin 2.0. Our method reduces average denoising steps by 60.2\%, 49.8\%, and 85.28\%, respectively, while maintaining baseline task success rates. In particular, our AnyStep training substantially improves model performance under a one-step denoising budget, increasing task success rates by 7.07\%, 12.08\%, and 8.94\% on Motus, FastWAM, and LingBotVA, respectively. Experiments on six real-world manipulation tasks further validate its effectiveness.
\end{abstract}

\section{Introduction}

World action models (WAMs)~\citep{bi2026motus,li2026causal,yuan2026fast} couple predictive visual modeling with action generation and have demonstrated strong performance in robotic manipulation. 
However, their repeated denoising through large diffusion transformers (DiTs) delays responses to new observations.
Reducing this iterative computation while preserving manipulation performance is therefore critical for efficient deployment.

\begin{figure}[t]
    \centering
    \includegraphics[
        width=0.95\linewidth,
        trim={10mm 107mm 10mm 10mm},
        clip
    ]{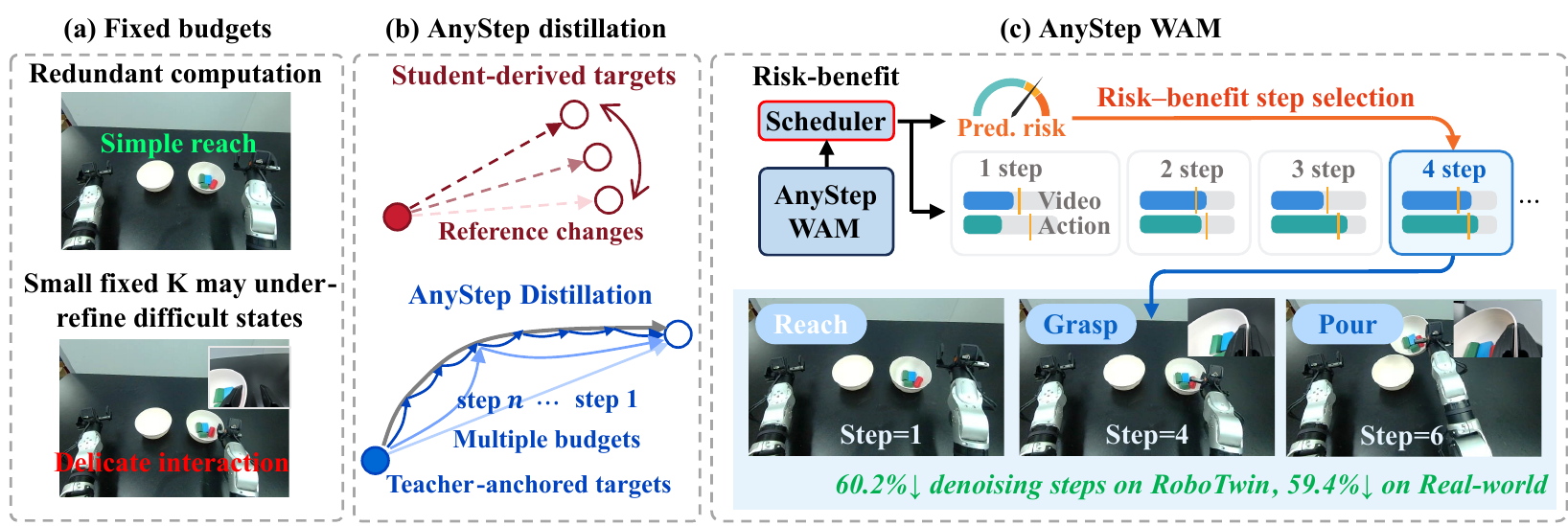}
    \caption{Overview of AnyStep WAM. (a) Fixed budgets can waste computation on easy states or under-refine difficult ones. (b) Teacher-anchored interval supervision enables prediction across budgets. (c) Risk-benefit scheduling adapts denoising steps to scene difficulty and predicted fidelity.}
    \label{fig:figure1}
    \vspace{-12pt}
\end{figure}

To reduce this cost, consistency learning~\citep{luo2023latent} and distribution matching distillation~\citep{yin2024one,yin2024improved} enable one- or few-step image and video generation.
Flash-WAM~\citep{akbari2026flash} extends consistency distillation to WAMs, accounting for distinct video and action noise regimes to enable one-step prediction.
However, fixed budgets cannot accommodate varying precision demands: some motions may need little refinement, while contact-sensitive actions may require more (Figure \ref{fig:figure1} (a)).
AnyFlow~\citep{gu2026anyflow} supports variable-step video generation, but its finite-difference supervision depends on the evolving student (Figure \ref{fig:figure1} (b)). In our WAM experiments, adapting AnyFlow produces highly variable targets and frequent gradient spikes during training, alongside low task success at one or two steps. The first challenge is therefore to train a single WAM that accurately predicts finite-interval action transitions across denoising budgets.

Accurate prediction across budgets leaves a separate question: \emph{how many} steps should the WAM use for the current action chunk? Adaptive methods in diffusion generation~\citep{zhang2025adadiff,le2026dsa} and robotic policies~\citep{hu2024adaflow,han2026adavla,ang2026adaptive,li2026elastic} vary computation with the input. For a WAM, changes in the teacher's denoising trajectory can indicate when coarse updates may be difficult, but they do not reveal how accurately the distilled student predicts at a particular budget.
Step selection must therefore consider both denoising difficulty and the student's budget-dependent fidelity. The second challenge is to estimate these quantities online and choose the smallest budget expected to provide sufficient fidelity.

As illustrated in Figure~\ref{fig:figure1} (c), we introduce \textbf{Any-Step World Action Model (AnyStep WAM)} for prediction across denoising budgets and adaptive inference. Its central training idea is \textit{budget-aligned integral flow-map distillation}: we sample intervals from candidate inference schedules and supervise each finite transition with the corresponding displacement along a frozen teacher trajectory. 
These targets supervise both long one-step and shorter multi-step updates without estimating temporal derivatives of the evolving student.
We also train corrective continuation from one-step student preview states, enabling preview reuse and refinement after budget selection.
Shared low-rank adapters support all candidate budgets within a single adapted WAM.

Building on this capability, we formulate adaptive inference as a \textbf{risk-benefit decision problem}.
Teacher-trajectory variation provides a difficulty proxy, while budget-specific student--teacher agreement estimates attainable fidelity.
From a single one-step preview, a lightweight scheduler predicts both quantities and selects the smallest budget whose predicted fidelity meets a difficulty-dependent threshold.
Reusing the preview in the selected trajectory avoids extra denoising passes, online teacher evaluation, and candidate rollouts.

We evaluate AnyStep WAM on Motus~\citep{bi2026motus}, LingBotVA~\citep{li2026causal}, and FastWAM~\citep{yuan2026fast} using RoboTwin~\citep{chen2025robotwin} and six real-world tasks. On RoboTwin, distillation improves one-step success by 7.07, 8.94, and 12.08 percentage points on Motus, LingBotVA, and FastWAM, respectively. Adaptive inference reduces their mean denoising steps by 60.2\%, 85.28\%, and 49.8\%, respectively, while keeping average success within 0.24 percentage points of the corresponding full-budget baselines. On the real robot, it achieves 1.67--6.14$\times$ per-call inference speedups with comparable or higher average success rates across the six tasks.

Our contributions are fourfold. \textbf{(i) Budget-aligned WAM distillation:} frozen-teacher interval targets and continuation training from student-generated states enable accurate prediction across multiple denoising budgets within one model. \textbf{(ii) Risk-benefit adaptive inference:} We estimate teacher-derived denoising difficulty and the student's fidelity at each budget, then adapt the step count to each interaction by selecting the smallest budget predicted to meet its fidelity requirement. \textbf{(iii) Efficient single-preview execution:} a one-step preview informs budget selection and is reused in the trajectory, avoiding an extra denoising pass, online teacher access, and candidate rollouts. \textbf{(iv) Experimental validation:} evaluations of three WAM backbones on RoboTwin and six real-world tasks show substantial reductions in denoising steps and latency while maintaining comparable or higher task success.

\section{Related Work}
\label{sec:related_work}

\noindent\textbf{Few- and any-step generation.}
Consistency learning~\citep{luo2023latent} and distribution matching distillation~\citep{yin2024one,yin2024improved} compress iterative diffusion or flow generation into one or a few steps, with related advances in robotic policies~\citep{yan2025maniflow,chen2026let,gao2026driftingvla} and joint video-action generation~\citep{akbari2026flash}. MeanFlow~\citep{geng2026mean} and AnyFlow~\citep{gu2026anyflow} further learn average velocities or flow maps over finite intervals for flexible-step generation. Unlike their derivative- or finite-difference-based supervision, we construct finite-interval targets from frozen teacher trajectories and align training intervals with candidate inference schedules for tunable-budget WAM distillation.

\noindent\textbf{Trajectory geometry and adaptive computation.}
Denoising trajectory geometry reflects both generative uncertainty and numerical difficulty. Concentrated posteriors yield more consistent velocities and straighter trajectories, whereas posterior dispersion and multimodality induce larger velocity variation and curvature~\citep{yang2026stable,rao2026geometryflowmatchinguncertaintycostfree}; highly curved trajectories also incur larger integration errors under coarse discretization~\citep{lee2023minimizing}.
These observations motivate adaptive computation according to denoising dynamics. Existing methods adjust denoising effort, network depth, or computation reuse using trajectory statistics, input-conditioned gating, or learned policies~\citep{hu2024adaflow,han2026adavla,zhang2025adadiff,yu2025d3p,li2026elastic,wang2026elastic,le2026dsa}. We use teacher-trajectory variation to characterize scene-dependent risk, while separately modeling the student's budget-dependent fidelity for computation allocation.

\section{Preliminaries}
\label{sec:preliminaries}

\subsection{World Action Models}
\label{sec:prelim_wam}
Given an observation $\vo$ and instruction $\vl$, WAMs incorporate
predictive visual modeling to generate an action chunk
$\va_{1:H}$ of horizon $H$~\citep{bi2026motus,li2026causal,yuan2026fast}.
We study their flow-based inference branches, which denoise
actions and, where applicable, future visuals, without assuming
a fixed video-action generation order.
Appendix~\ref{app:wam_formulation} details the different formulations.

\subsection{Flow Matching and Flow Maps}
\label{sec:prelim_flow}

\textbf{Flow matching.}
Flow matching~\citep{lipman2023flowmatchinggenerativemodeling}
transports noise $\vx_1$ to data $\vx_0$ by solving
$\mathrm{d}\vx_t/\mathrm{d}t=\vv(\vx_t,t;\vc)$
from $t=1$ to $0$, where $\vv$ is the
velocity field conditioned on context $\vc$.

\textbf{Flow maps.}
A flow map instead represents a finite-time transition along the
ODE trajectory. For $0\leq r<t\leq1$, its interval-average
velocity is
\begin{equation}
    \bar{\vv}(\vx_t,t,r;\vc)
    = \frac{1}{t-r}
      \int_r^t \vv(\vx_s,s;\vc)\,\mathrm{d}s.
    \label{eq:avg_velocity}
\end{equation}
A neural approximation $\vu_{\vtheta}$ parameterizes
the transition as
\begin{equation}
    f_{\vtheta}(\vx_t,t,r;\vc)
    = \vx_t-(t-r)\vu_{\vtheta}(\vx_t,t,r;\vc).
    \label{eq:flow_map}
\end{equation}
Conditioning on both endpoints supports different transition
intervals and inference budgets.

\textbf{Derivative-based flow-map supervision.}
MeanFlow~\citep{geng2026mean} constructs a target relating
average and instantaneous velocities:
\begin{equation}
    \vu_{\mathrm{tgt}}
    = \vv(\vx_t,t;\vc)
      -(t-r)\frac{\mathrm{d}}{\mathrm{d}t}
      \vu_{\vtheta}(\vx_t,t,r;\vc),
    \label{eq:meanflow_identity}
\end{equation}
and optimizes
\begin{equation}
    \mathcal{L}_{\mathrm{MF}}
    = \mathbb{E}\left[
        \left\|
        \vu_{\vtheta}(\vx_t,t,r;\vc)
        -\operatorname{sg}(\vu_{\mathrm{tgt}})
        \right\|_2^2
      \right],
    \label{eq:meanflow_loss}
\end{equation}
where $\operatorname{sg}$ denotes stop-gradient. The total derivative follows the ODE trajectory with $r$ fixed. MeanFlow computes it through a Jacobian-vector product, whereas AnyFlow~\citep{gu2026anyflow} uses central finite differences. These targets depend on the evolving student and can be sensitive to prediction and optimization noise~\citep{geng2026improvedmeanflowschallenges,Tu_2026_CVPR,gu2026anyflow}. Appendix~\ref{app:appendix_prelim_flow} provides the full objectives and further discussion.

\section{Method}
\label{sec:method}

\begin{figure}[t]
    \vspace{-10pt}
    \centering
    \includegraphics[width=\linewidth, trim={12mm 108mm 6mm 10mm}, clip]{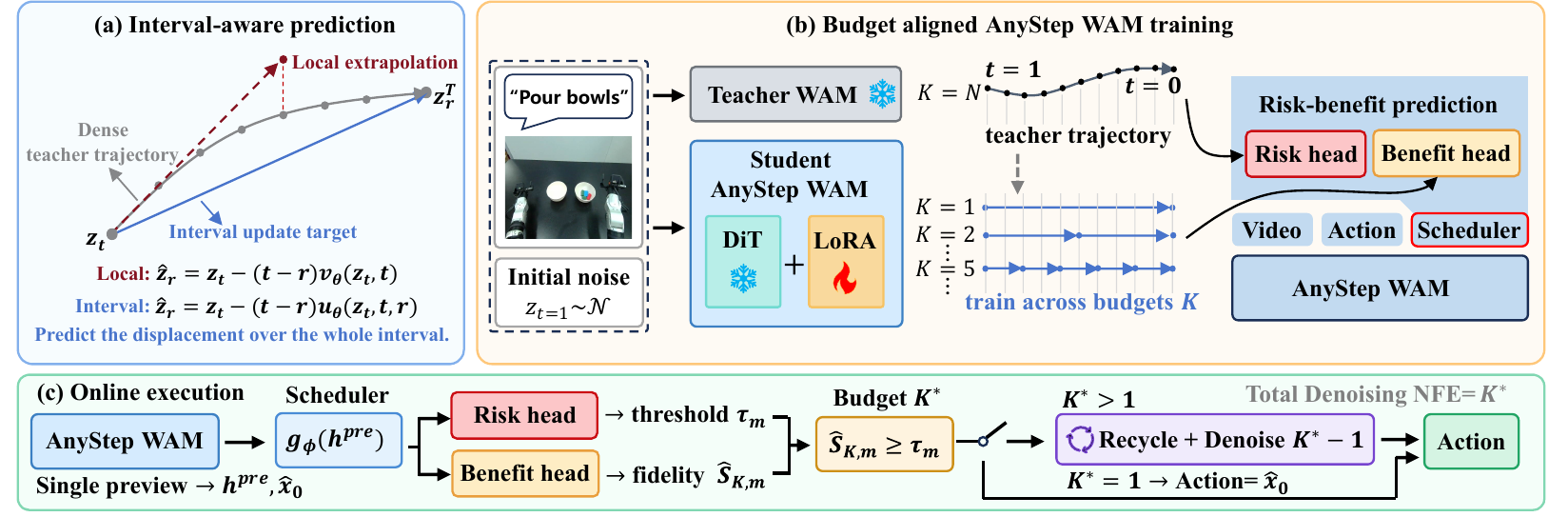}
    \vspace{-10pt}
    \caption{\textbf{Budget-aligned distillation and adaptive inference.} (a) Teacher-trajectory displacements supervise finite-interval updates. (b) Shared LoRA adapters support multiple budgets, while teacher dynamics and student-teacher agreement provide offline risk and benefit targets. (c) Single-preview scheduling and recycling require exactly $K^\star$ denoising evaluations, without online teacher access or candidate rollouts.}
    \label{fig:framework}
    \vspace{-10pt}
\end{figure}

Our framework (Figure~\ref{fig:framework}) combines an AnyStep WAM for tunable-budget generation with a lightweight risk-benefit scheduler. We first adapt a pretrained WAM using integral-supervised flow maps, then train the scheduler on offline teacher-student evaluations to predict chunk difficulty and budget-dependent fidelity. At inference, a single preview supplies features for budget selection and is reused in the selected trajectory. Here, $m\in\mathcal M$ indexes denoised modalities, with $\mathcal M=\{\mathrm z,\mathrm a\}$ for joint video-action models and $\mathcal M=\{\mathrm a\}$ for action-only inference.

\subsection{Enabling AnyStep WAMs with Integral Flow-Map Distillation}
\label{sec:any_step_distillation}
We adapt the denoising branches of a pretrained WAM to the flow-map parameterization in Eq.~\ref{eq:flow_map}, enabling a shared model to predict finite-time transitions under different denoising budgets. We train these transitions using explicit teacher-integrated targets rather than derivative-based supervision.

\textbf{Integral supervision from teacher trajectories.}
Given a condition $\vc=(\vo,\vl)$ and initial noise $\vx_1=\{\vx_1^m, m\in \mathcal M\}$, we first run the frozen teacher WAM using a fine-grained reference solver. Let $\vx_t^{\mathrm T,m}$ denote the teacher state of modality $m$ at denoising time $t$, with $\vx_1^{\mathrm T, m}=\vx_1^m$. For two states on the same trajectory with $r<t$, we define the teacher average velocity of modality $m$ over the interval $[r,t]$ as
\begin{equation}
    \overline{\vv}^{\mathrm{T},m}_{t,r}
    =
    \frac{\vx_t^{\mathrm{T},m}-\vx_r^{\mathrm{T},m}}{t-r},
    \qquad r<t.
    \label{eq:teacher_avg_vel}
\end{equation}
This finite-displacement target directly approximates the integral average velocity in Eq.~\ref{eq:avg_velocity}. Importantly, because both endpoints are determined by the frozen teacher, the target is independent of the student parameters and does not require derivative-based supervision through the student dynamics.

\textbf{Budget-aligned interval sampling.}
We consider a set of candidate inference budgets $\mathcal K$. Each budget $K\in\mathcal K$ is associated with a fixed inference schedule $\mathcal S_K=\left\{1=t_0^{(K)}>t_1^{(K)}>\cdots>t_K^{(K)}=0 \right\}$.
Since different budgets induce different transition intervals, we align training with the transitions that the model will encounter at inference. Specifically, we first sample a budget $K\in\mathcal K$ and then an interval
index $i\in\{0,\ldots,K-1\}$, yielding $(t,r)=(t_i^{(K)},t_{i+1}^{(K)})$. The corresponding teacher state $\vx_t^{\mathrm T}$ and average velocity $\overline{\vv}_{t,r}^{\mathrm T}$ provide supervision for this transition, exposing the student to the finite-time intervals used by the candidate inference schedules.

\textbf{Training objective.}
We train the student on these budget-aligned intervals and complement this supervision with local, endpoint, and recovery terms. All four terms share a dimension-normalized regression form.
For an input state $\vx_t=\{\vx_t^m\}_{m\in\mathcal M}$ and target velocities $\vv^\star=\{\vv^{\star,m}\}_{m\in\mathcal M}$, we define
\begin{equation}
\mathcal R(\vx_t,t,r,\vc;\vv^\star)
=
\sum_{m\in\mathcal M}
\frac{1}{d_m}
\left\|
\vu_{\vtheta}^{m}(\vx_t,t,r;\vc)
-
\vv^{\star,m}
\right\|_2^2,
\label{eq:interval_regression_loss}
\end{equation}
where $\vu_{\vtheta}^{m}$ denotes the modality-$m$ output conditioned on the joint input state, and $d_m$ is its output dimensionality. The normalization removes the direct dependence of each modality's loss contribution on its dimensionality.
To align training with the preview-recycled inference procedure described in Section~\ref{sec:adaptive_inference}, the recovery loss uses a detached student-generated state $\bar{\vx}_t=\operatorname{sg}(\widetilde{\vx}_t^{(K)})$.
For each $K>1$, we restart the frozen teacher once from the detached first recycled state and integrate to time zero, obtaining a budget-specific reference trajectory $\vx_s^{\mathrm T,K}$ for all remaining intervals.
We define the corrective target $\vv_{t,r}^{\mathrm{rec}}=(\bar{\vx}_t-\vx_r^{\mathrm T,K})/(t-r)$ using this restarted reference, with no gradient through the target.
Using this regression form, the complete training objective is
\begin{align}
\mathcal L_{\mathrm{WAM}}
&=
\lambda_{\mathrm{int}}
\underbrace{
\mathbb E_{\vc,\vx_1,K,i}
\left[
\mathcal R(
\vx_t^{\mathrm T},t,r,\vc;
\overline{\vv}_{t,r}^{\mathrm T}
)
\right]
}_{\mathcal L_{\mathrm{int}}}
+
\lambda_{\mathrm{FM}}
\underbrace{
\mathbb E_{\vc,\vx_0,\vx_1,t}
\left[
\mathcal R(
\vx_t,t,t,\vc;\vx_1-\vx_0
)
\right]
}_{\mathcal L_{\mathrm{FM}}}
\nonumber\\
&+
\lambda_{\mathrm{end}}
\underbrace{
\mathbb E_{\vc,\vx_1,t}
\left[
\mathcal R(
\vx_t^{\mathrm T},t,0,\vc;
\overline{\vv}_{t,0}^{\mathrm T}
)
\right]
}_{\mathcal L_{\mathrm{end}}}
+
\lambda_{\mathrm{rec}}
\underbrace{
\mathbb E_{\vc,\vx_1,K,i}
\left[
\mathcal R(
\bar{\vx}_t,t,r,\vc;
\vv_{t,r}^{\mathrm{rec}}
)
\right]
}_{\mathcal L_{\mathrm{rec}}},
\label{eq:wam_objective}
\end{align}
where $(t,r)=(t_i^{(K)},t_{i+1}^{(K)})$ in $\mathcal L_{\mathrm{int}}$.
This $\mathcal L_{\mathrm{int}}$ term learns the finite-time transitions specified by the candidate schedules.
The local term $\mathcal L_{\mathrm{FM}}$ preserves instantaneous-velocity prediction using standard flow-matching targets with identical time inputs, $r=t$.
The endpoint term $\mathcal L_{\mathrm{end}}$ supervises transitions to $r=0$, with dedicated $(t,r)=(1,0)$ supervision for the one-step preview.
Finally, $\mathcal L_{\mathrm{rec}}$ trains continuation from detached states along preview-recycled student trajectories.
For each $K>1$, it averages over the remaining intervals
$(t,r)=(t_i^{(K)},t_{i+1}^{(K)})$, $i=1,\ldots,K-1$.
The corrective target accounts for deviations from the restarted reference, so exact matching reaches its next endpoint $\vx_r^{\mathrm T,K}$.
This additional supervision aligns the student with the preview-reuse procedure in Section~\ref{sec:adaptive_inference}.

We implement this adaptation using LoRA~\citep{hu2021lora} in selected attention, feed-forward, and time-conditioning projections. All pretrained weights remain frozen, and the same adapters are shared across budgets. For models that denoise only actions at inference~\citep{yuan2026fast}, we retain only the action losses in our distillation objective.

\subsection{Self-Aware Risk-Benefit Prediction}
\label{sec:self_aware_scheduler}

Different interaction contexts may require different amounts of denoising computation. To guide budget selection without evaluating multiple candidates online, we train a lightweight predictor to estimate teacher-trajectory difficulty and budget-specific student fidelity from offline supervision.

\textbf{Risk supervision.}
We use velocity variation along the original teacher trajectory initialized at $\vx_1$ as a budget-independent proxy for denoising difficulty. Let $\vv_{j,m}^{\mathrm T}$ denote the modality-$m$ teacher velocity at the $j$th reference-solver step, with $N$ steps in total. We measure directional disagreement and relative velocity change:
\begin{equation}
\begin{aligned}
    \kappa_m^{\mathrm{dir}}
    =
    \frac{1}{N-1}
    \sum_{j=1}^{N-1}
    \left[
        1-\operatorname{cos}
        \left(\vv_{j+1,m}^{\mathrm T},\vv_{j,m}^{\mathrm T}\right)
    \right], 
    \  \kappa_m^{\mathrm{rel}}
    =
    \frac{1}{N-1}
    \sum_{j=1}^{N-1}
    \frac{
        \operatorname{RMS}
        \left(\vv_{j+1,m}^{\mathrm T}-\vv_{j,m}^{\mathrm T}\right)
    }{
        \max\left\{
            \operatorname{RMS}\left(\vv_{j,m}^{\mathrm T}\right),
            \epsilon
        \right\}
    }.
\end{aligned}
\label{eq:teacher_risk_statistics}
\end{equation}
Here, cosine similarity and root-mean-square (RMS) are computed over valid generated coordinates, with numerical safeguards $\epsilon>0$. For each modality, we average the training-set percentile ranks of the two statistics to obtain $\rho_m=\frac{1}{2}[F_m^{\mathrm{dir}}(\kappa_m^{\mathrm{dir}}) +F_m^{\mathrm{rel}}(\kappa_m^{\mathrm{rel}})]$, where $F_m^{\mathrm{dir}}$ and $F_m^{\mathrm{rel}}$ are empirical percentile mappings to $[0,1]$. The resulting $\rho_m$ represents relative denoising difficulty rather than task-failure probability.

\textbf{Budget benefit supervision.}
Risk characterizes teacher-trajectory variability but does not
directly reflect the student's approximation error at a given budget.
For each candidate budget $K\in\mathcal K$, we evaluate offline
the same preview-recycled denoising path used at deployment.
Let $\widetilde{\vx}_{t_i^{(K)}}^{(K)}$ denote the student state
along this path.
The first interval uses the preview velocity
$\vv_{K,0}^{\mathrm{exec},m}
=\vu_{\vtheta}^{m}(\vx_1,1,0;\vc)$,
while subsequent intervals use
$\vv_{K,i}^{\mathrm{exec},m}
=\vu_{\vtheta}^{m}(
\widetilde{\vx}_{t_i^{(K)}}^{(K)},
t_i^{(K)},t_{i+1}^{(K)};\vc)$,
$i=1,\ldots,K-1$.
Preview recycling scales the displacement by $1-t_1^{(K)}$, so the average student velocity over the first interval remains the shared preview velocity.
We compare it with $\vv_{K,0}^{\mathrm{rec},m}=\overline{\vv}_{1,t_1^{(K)}}^{\mathrm T,m}$ over the same interval along the original teacher trajectory.
For $i\geq1$, $\vv_{K,i}^{\mathrm{rec},m}$ is the modality-$m$ corrective target defined above, evaluated at $(t,r)=(t_i^{(K)},t_{i+1}^{(K)})$ using the teacher trajectory restarted once from the recycled state.
For $K=1$, only the first interval applies, with teacher endpoint $\vx_0^{\mathrm T}$.

\begin{equation}
S_{K,m}
=
\sum_{i=0}^{K-1}w_i^{(K)}
\psi\left(
\vv_{K,i}^{\mathrm{exec},m},
\vv_{K,i}^{\mathrm{rec},m}
\right),
\qquad
\sum_{i=0}^{K-1}w_i^{(K)}=1.
\label{eq:budget_benefit}
\end{equation}

Here, $\psi$ combines directional and normalized velocity agreement:

\begin{equation*}
\psi(\mathbf p,\mathbf q)
=
\frac{1+\operatorname{cos}(\mathbf p,\mathbf q)}{4}
+
\frac{1}{2}
\exp\left(
-\frac{
2\operatorname{RMS}(\mathbf p-\mathbf q)
}{
\max\left\{
\operatorname{RMS}(\mathbf p)
+\operatorname{RMS}(\mathbf q),
\epsilon
\right\}
}
\right).
\end{equation*}

The weights
$w_i^{(K)}=2^{-i}/\sum_{j=0}^{K-1}2^{-j}$,
$i=0,\ldots,K-1$, emphasize earlier denoising intervals.
Thus, $S_{K,m}\in[0,1]$ summarizes preview agreement and subsequent recovery fidelity. It remains a proxy rather than a direct measure of terminal action error or task success.

\textbf{Learning from one-step previews.}
For the same condition and initial noise used to construct
the supervision targets, a one-step evaluation
$\vu_{\vtheta}(\vx_1,1,0;\vc)$ provides hidden features and
predicted velocities.
Their summaries, together with contextual features, form
the scheduler input $\mathbf h^{\mathrm{pre}}$.
A lightweight predictor $g_{\phi}$ produces
$g_{\phi}(\mathbf h^{\mathrm{pre}})
=\bigl(\{\widehat{\rho}_m\}_{m\in\mathcal M},
\{\widehat S_{K,m}\}_{K\in\mathcal K,m\in\mathcal M}\bigr)$.
Both prediction heads use sigmoid outputs and are supervised with Smooth L1 losses against the corresponding offline targets. 
Architectural details and training configurations are provided
in Appendix~\ref{appendix:implementation}.

\subsection{Risk-Conditioned Adaptive Inference}
\label{sec:adaptive_inference}

Adaptive inference uses predicted risk and budget-specific fidelity to select the smallest budget meeting risk-adaptive predicted requirements, then recycles the one-step preview to initialize the selected sampling schedule without candidate rollouts.

\textbf{Risk-conditioned budget selection.}
At inference, a single evaluation produces the preview velocity $\vu^{\mathrm{pre}}=\vu_{\vtheta}(\vx_1,1,0;\vc)$, the prediction $\widehat{\vx}_0=\vx_1-\vu^{\mathrm{pre}}$, and the features used by the risk-benefit predictor. 
For each active modality $m$, predicted risk interpolates
between the easy and hard fidelity thresholds
$\tau_m^{\mathrm{easy}}$ and $\tau_m^{\mathrm{hard}}$:
\begin{equation}
    \tau_m
    =
    \tau_m^{\mathrm{easy}}
    +
    \left(\tau_m^{\mathrm{hard}}-\tau_m^{\mathrm{easy}}\right)
    \widehat{\rho}_m
    \label{eq:risk_threshold}
\end{equation}
Higher predicted difficulty therefore imposes a stricter fidelity requirement. We select the smallest budget $K^\star$ satisfying $\widehat S_{K^\star,m}\geq\tau_m$ for all active modalities, or the budget with the largest summed predicted benefit if none qualifies.

\textbf{Preview recycling.}
As illustrated in Figure~\ref{fig:framework}(c), the preview is reused within the selected budget. If $K^\star=1$, we return $\widehat{\vx}_0$ directly. Otherwise, we re-noise it to the first intermediate time using the original noise:
$\widetilde{\vx}_{t_1^{(K^\star)}}^{(K^\star)}
=t_1^{(K^\star)}\vx_1
+(1-t_1^{(K^\star)})\widehat{\vx}_0$.
We then perform the remaining $K^\star-1$ interval updates, requiring exactly $K^\star$ denoising evaluations including the preview, without online teacher access or candidate rollouts. The scheduler adds only lightweight prediction overhead.

\section{Experiment}

\textbf{Benchmarks.}
We evaluate Motus~\citep{bi2026motus}, FastWAM~\citep{yuan2026fast}, and LingBotVA~\citep{li2026causal} through comparisons, ablations, and analyses.
RoboTwin 2.0~\citep{chen2025robotwin} provides 50 bimanual manipulation tasks under clean and randomized settings, with randomization covering clutter, lighting, background textures, and tabletop heights. Real-world evaluation uses a Unitree G1D dual-arm robot with Unitree Dex1-1 grippers on six tasks: Bowl Pouring, Clean Table, Put Block, Put Cup, Stack Blocks, and Stack Bowls.

\textbf{Implementation and evaluation.}
Each backbone is trained on four NVIDIA H100 GPUs for 8,000 steps with a global batch size of 32. RoboTwin evaluation uses a single NVIDIA A100 with 100 trials per task for every method and ablation variant; real-world inference runs locally on a single NVIDIA RTX 5090 with 20 trials per task on average. More details are in Appendix~\ref{appendix:implementation}. For budget selection, easy/hard thresholds use the 10th/20th percentiles of offline training-set video similarity and the 20th/80th percentiles of action similarity; Section~\ref{sec:ablation_experiments} evaluates threshold sensitivity.

\subsection{Simulation experiments}
\label{simulation_experiments}

\begin{figure}[h]
    \centering
    \includegraphics[width=0.95\linewidth, trim={10mm 69mm 10mm 12mm}, clip]{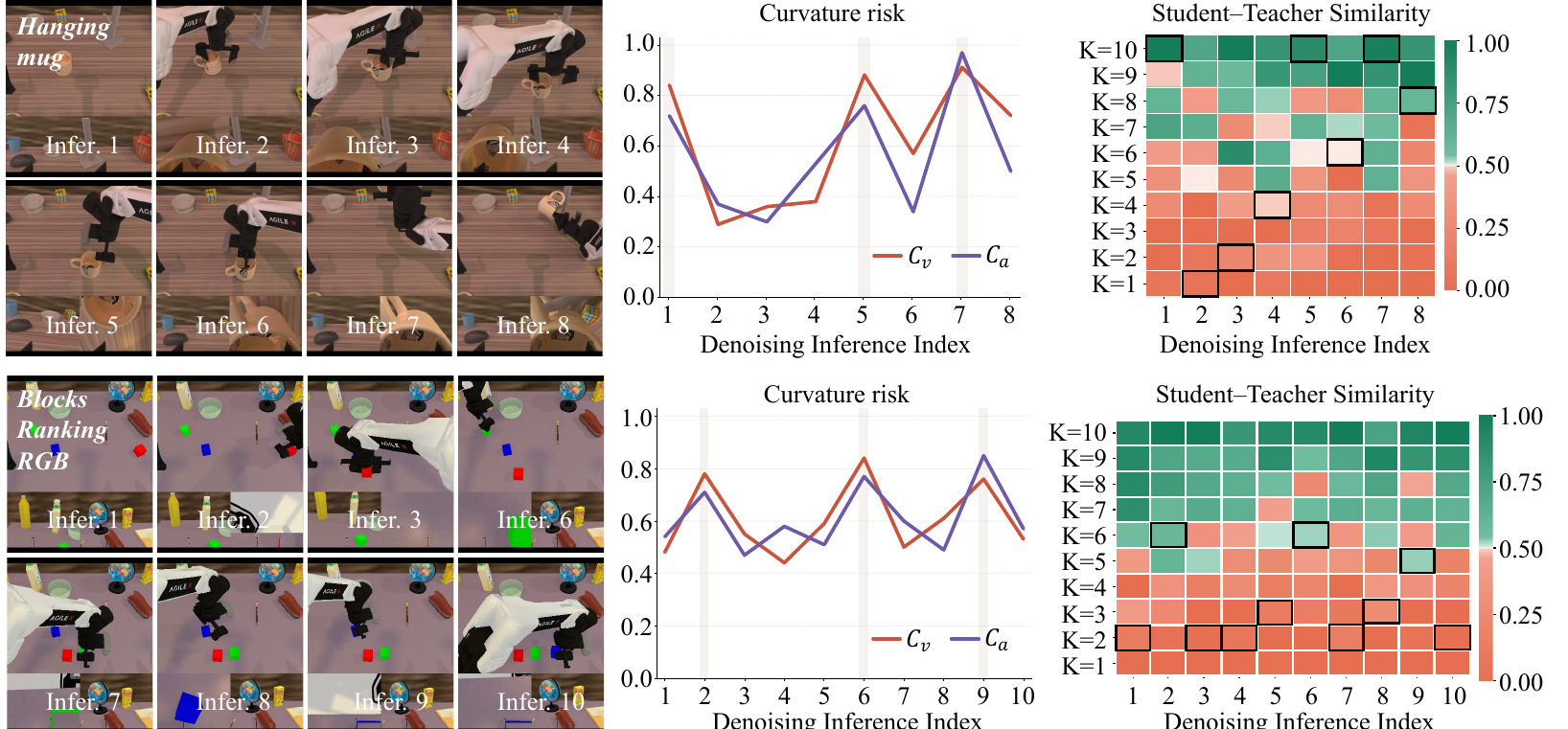}
    \caption{Adaptive denoising rollouts. The left column shows representative observations. The middle column plots predicted video ($C_v$) and action ($C_a$) curvature risks, shaded regions marking high-risk stages. The right column shows student-teacher velocity similarities, combined by their minimum. Black boxes mark selected budgets.}
    \label{fig:robotwin_demo}
\end{figure}

\begin{table}[h]
\caption{RoboTwin 2.0 results. Base denotes the original model. SR denotes success rate (\%); Steps denotes mean denoising steps for adaptive methods. Per-task results are in Appendix~\ref{appendix:all_tasks_results}.}
\label{tab:robotwin_adaptive}
\begin{center}
\normalfont\footnotesize
\setlength{\tabcolsep}{0pt}
\renewcommand{\arraystretch}{1.0}
\begin{adjustbox}{max width=\linewidth}
\begin{tabular}{@{}l@{\hspace{0.6em}}
*{6}{>{\centering\arraybackslash}m{2.5em}}
@{\hspace{1em}}
*{6}{>{\centering\arraybackslash}m{2.5em}}
@{\hspace{1em}}
*{6}{>{\centering\arraybackslash}m{2.5em}}@{}}
\toprule
\multirow{3}{*}{Method}
& \multicolumn{6}{c}{Motus~\citep{bi2026motus}}
& \multicolumn{6}{c}{FastWAM~\citep{yuan2026fast}}
& \multicolumn{6}{c}{LingBotVA~\citep{li2026causal}} \\
\cmidrule(lr){2-7}\cmidrule(lr){8-13}\cmidrule(lr){14-19}
& \multicolumn{2}{c}{Clean}
& \multicolumn{2}{c}{Random}
& \multicolumn{2}{c}{Average}
& \multicolumn{2}{c}{Clean}
& \multicolumn{2}{c}{Random}
& \multicolumn{2}{c}{Average}
& \multicolumn{2}{c}{Clean}
& \multicolumn{2}{c}{Random}
& \multicolumn{2}{c}{Average} \\
\cmidrule(lr){2-3}\cmidrule(lr){4-5}\cmidrule(lr){6-7}
\cmidrule(lr){8-9}\cmidrule(lr){10-11}\cmidrule(lr){12-13}
\cmidrule(lr){14-15}\cmidrule(lr){16-17}\cmidrule(lr){18-19}
& SR & Steps & SR & Steps & SR & Steps
& SR & Steps & SR & Steps & SR & Steps
& SR & Steps & SR & Steps & SR & Steps \\
\midrule
Base
& 76.00 & 1 & 74.10 & 1 & 75.05 & 1
& 77.26 & 1 & 77.14 & 1 & 77.20 & 1
& 74.36 & 1 & 73.18 & 1 & 73.77 & 1 \\
Flash-WAM
& 79.20 & 1 & 78.66 & 1 & 78.93 & 1
& 82.26 & 1 & 81.14 & 1 & 81.70 & 1
& 82.56 & 1 & 80.26 & 1 & 81.41 & 1 \\
\textbf{AnyStep (Ours)}
& \textbf{84.14} & 1 & \textbf{80.10} & 1 & \textbf{82.12} & 1
& \textbf{89.12} & 1 & \textbf{89.44} & 1 & \textbf{89.28} & 1
& \textbf{83.80} & 1 & \textbf{81.62} & 1 & \textbf{82.71} & 1 \\
\midrule
Base
& \textbf{88.66} & 10 & 87.02 & 10 & 87.84 & 10
& 91.88 & 10 & \textbf{91.78} & 10 & \textbf{91.83} & 10
& \textbf{92.93} & 25/50 & 91.55 & 25/50 & \textbf{92.24} & 25/50 \\
AnyStep+Gate
& 87.60 & 5.23 & 86.92 & 4.95 & 87.26 & 5.09
& 90.28 & 6.34 & 90.04 & 6.98 & 90.16 & 6.66
& 90.16 & 4.58 & 90.52 & 5.04 & 90.34 & 4.81 \\
\textbf{AnyStep (Ours)}
& 88.12 & \textbf{4.02} & \textbf{87.80} & \textbf{3.94} & \textbf{87.96} & \textbf{3.98}
& \textbf{92.12} & \textbf{5.02} & 91.06 & \textbf{5.03} & 91.59 & \textbf{5.03}
& 92.74 & \textbf{3.54} & \textbf{91.70} & \textbf{3.82} & 92.22 & \textbf{3.68} \\
\bottomrule
\end{tabular}
\end{adjustbox}
\end{center}
\end{table}

\textbf{Results on RoboTwin.}
In Figure~\ref{fig:robotwin_demo}, contact-intensive stages tend to exhibit higher predicted risk, favoring larger denoising budgets that meet stricter fidelity requirements.
Table~\ref{tab:robotwin_adaptive} compares task success rates and generation efficiency across three WAM backbones. Base Motus and FastWAM use a fixed 10-step budget, whereas the serial LingBotVA architecture uses 25 steps for its video DiT and 50 for its action DiT. \textbf{Ours} maintains average SR within 0.24 percentage points of the baselines while selecting mean budgets of 3.98, 5.03, and 3.68 steps, respectively. These correspond to reductions of 60.2\% for Motus, 49.8\% for FastWAM, and 85.28\%/92.64\% for LingBotVA's video/action branches. 
On a single NVIDIA A100 GPU, average per-inference-call latency drops from 1.932 to 0.859\,s for Motus, 0.496 to 0.295\,s for FastWAM, and 9.732 to 3.247\,s for LingBotVA, with respective budget-selection module overheads of only 2.881, 2.4996, and 2.9772\,ms per prediction.

To the best of our knowledge, prior work has not specifically addressed adaptive budget selection for flow-map-trained WAMs. We therefore construct \textbf{AnyStep+Gate}, which applies training-free percentile gating to action-coordination and visual-temporal scores on the same distilled models (Appendix~\ref{appendix:Training-free gating baseline}). Our scheduler improves average SR over gating by 0.70, 1.43, and 1.88 percentage points while reducing average denoising steps by 21.8\%, 24.5\%, and 23.5\% on Motus, FastWAM, and LingBotVA, respectively, supporting fidelity-aware allocation. Under a matched one-step denoising budget, \textbf{AnyStep 1 step} improves average SR over \textbf{Base 1 step} by 7.07, 12.08, and 8.94 percentage points on Motus, FastWAM, and LingBotVA, respectively. It also outperforms \textbf{Flash-WAM}~\citep{akbari2026flash} by 3.19, 7.58, and 1.30 percentage points, respectively, supporting the effectiveness of our teacher-trajectory distillation for one-step action prediction across architectures. 

\begin{figure}[h]
    \centering
    \includegraphics[width=1\linewidth, trim={0mm 0mm 0mm 0mm}, clip]{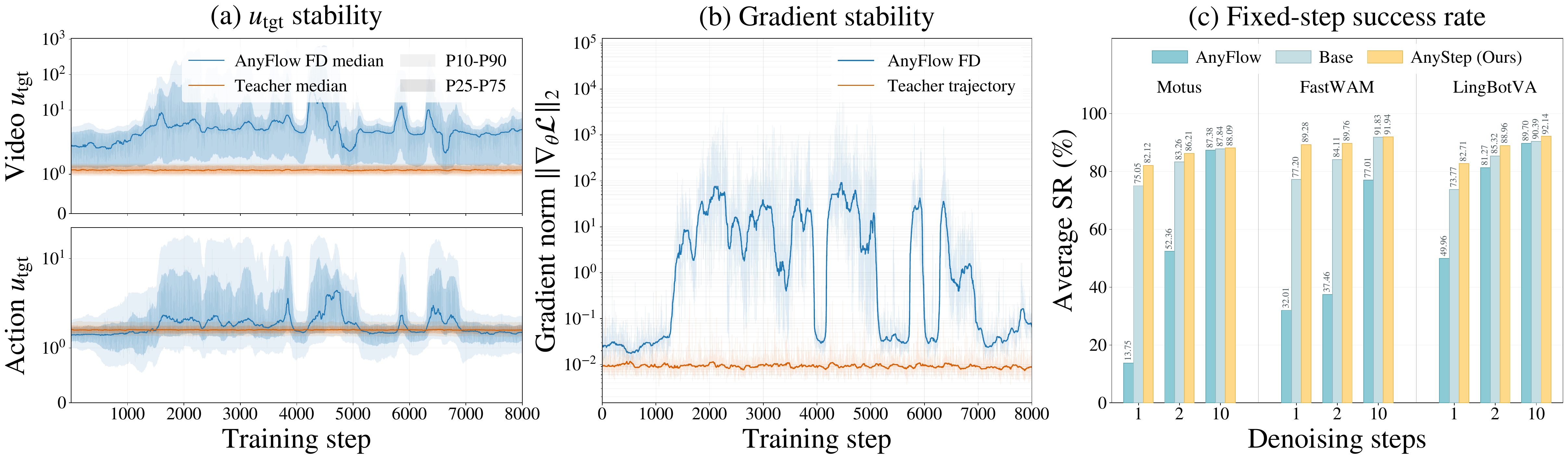}
    \vspace{-12pt}
    \caption{Training stability and fixed-budget performance. (a) Per-sample target RMS distributions. (b) Gradient norms.  (c) Average success rates at fixed denoising steps.}
    \label{fig:fdvsteacher}
    \vspace{-10pt}
\end{figure}

\textbf{Fixed-budget generation and training stability.} Figure~\ref{fig:fdvsteacher} compares finite-difference and teacher-trajectory supervision. Finite-difference targets exhibit wider sample-level distributions and produce frequent gradient spikes, with $45.9\%$ of steps exceeding the clipping threshold of $1.0$. Frozen-teacher targets remain concentrated and yield substantially more stable gradients (Fig.~\ref{fig:fdvsteacher}(a-b)). This stability also improves fixed-budget performance (Fig.~\ref{fig:fdvsteacher}(c)). At one denoising step, AnyStep improves average SR over Base by $7.07$,
$12.08$, and $8.94$ points on Motus, FastWAM, and LingBotVA, respectively, and remains best at 2 and 10 steps. AnyStep therefore provides stable budget-aligned supervision without requiring teacher evaluation at inference. Results for additional denoising budgets are provided in Appendix~\ref{appendix:fixed_denoise_step_results}.

\subsection{Real-world experiments}
\label{real_world_experiments}

\begin{figure}[h]
    \centering
    \includegraphics[width=\linewidth, trim={8mm 55mm 8mm 40mm}, clip]{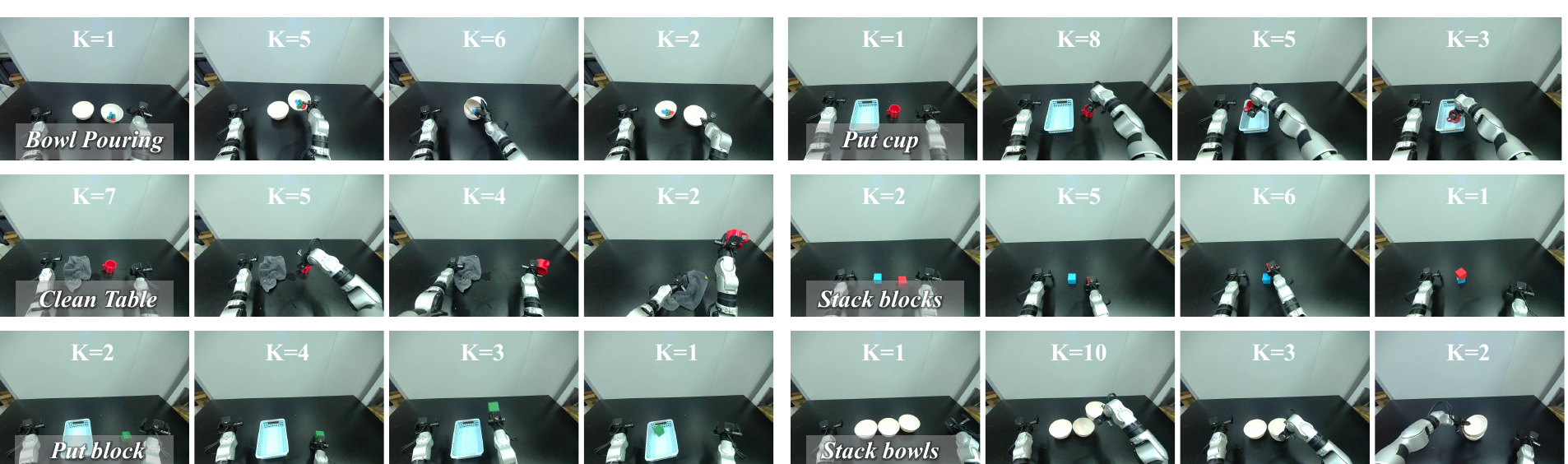}
    \vspace{-8pt}
    \caption{Real-world rollouts across six tasks on a Unitree G1D dual-arm robot. $K$ denotes the denoising budget adaptively selected for each shown inference call.}
    \label{fig:real_world}
    \vspace{-10pt}
\end{figure}

\begin{table}[h]
\vspace{-8pt}
\caption{Real-world results on six manipulation tasks.}
\label{tab:realworld}
\begin{center}
\normalfont
\setlength{\tabcolsep}{2pt}
\begin{adjustbox}{max width=0.9\linewidth}
\begin{tabular}{@{}l*{12}{c}@{}}
\toprule
\multirow{2}{*}{Task}
& \multicolumn{2}{c}{Motus}
& \multicolumn{2}{c}{AnyStep-Motus}
& \multicolumn{2}{c}{FastWAM}
& \multicolumn{2}{c}{AnyStep-FastWAM}
& \multicolumn{2}{c}{LingBotVA}
& \multicolumn{2}{c}{AnyStep-LingBotVA} \\
\cmidrule(lr){2-3}\cmidrule(lr){4-5}
\cmidrule(lr){6-7}\cmidrule(lr){8-9}
\cmidrule(lr){10-11}\cmidrule(lr){12-13}
& SR & Steps & SR & Steps
& SR & Steps & SR & Steps
& SR & Steps & SR & Steps \\
\midrule
Bowl Pouring
& 65 & 10 & 65 & 4.13
& 70 & 10 & 75 & 5.20
& 70 & 25/50 & 70 & 3.72 \\
Clean Table
& 60 & 10 & 60 & 5.75
& 65 & 10 & 60 & 2.21
& 55 & 25/50 & 60 & 5.06 \\
Put Block
& 70 & 10 & 75 & 2.56
& 75 & 10 & 70 & 3.54
& 75 & 25/50 & 70 & 2.21 \\
Put Cup
& 70 & 10 & 75 & 3.95
& 75 & 10 & 85 & 2.31
& 70 & 25/50 & 75 & 3.47 \\
Stack Blocks
& 55 & 10 & 60 & 2.81
& 50 & 10 & 55 & 4.62
& 60 & 25/50 & 60 & 4.45 \\
Stack Bowls
& 75 & 10 & 80 & 5.17
& 85 & 10 & 85 & 3.77
& 80 & 25/50 & 85 & 2.31 \\
\midrule
Avg.
& 65.83 & 10 & \textbf{69.17} & \textbf{4.06}
& 70.00 & 10 & \textbf{71.67} & \textbf{3.61}
& 68.33 & 25/50 & \textbf{70.00} & \textbf{3.54} \\
\bottomrule
\end{tabular}
\end{adjustbox}
\end{center}
\end{table}
 
We evaluate six real-world manipulation tasks on a Unitree G1D dual-arm robot. Each base model and its AnyStep variant are trained using 50 episodes per task for 8,000 optimization steps on four NVIDIA H100 GPUs with a global batch size of 32. Figure~\ref{fig:real_world} shows stage-dependent budgets, with more steps allocated to grasping and pouring in Bowl Pouring, illustrating how adaptive scheduling uses AnyStep's tunable-budget capability to refine demanding interactions. Table~\ref{tab:realworld} shows average SR increasing to 69.17\%, 71.67\%, and 70.00\% on Motus, FastWAM, and LingBotVA, while mean denoising steps decrease by 59.4\%, 63.9\%, and 85.84\%, respectively. On a single NVIDIA RTX 5090 GPU, average per-inference-call latency drops from 1.868 to 1.117\,s for Motus, 0.280 to 0.122\,s for FastWAM, and 4.654 to 0.758\,s for LingBotVA, yielding respective speedups of $1.67\times$, $2.30\times$, and $6.14\times$. These results demonstrate faster inference with higher average task success.

\subsection{Ablation experiments}
\label{sec:ablation_experiments}

Table~\ref{tab:ablation} reports fixed-step and adaptive results for loss ablations, together with scheduler generalization and threshold sensitivity on the Motus backbone. Additional ablations on FastWAM and LingBotVA are reported in Appendix~\ref{app:additional_ablation_backbones}.
The \textbf{w/o $\boldsymbol{\mathcal{L}_{\mathrm{int}}}$} variant removes budget-aligned teacher-interval supervision. Compared with \textbf{Full (Ours)}, its average one-step SR decreases by 1.40 percentage points, while two-step SR drops from 86.21\% to 78.94\% and adaptive SR from 87.96\% to 78.09\%, increasing the average budget from 3.98 to 5.27 steps. This highlights the importance of interval supervision for accurate tunable-budget prediction.
The \textbf{w/o $\boldsymbol{\mathcal{L}_{\mathrm{end}}}$} variant removes dedicated endpoint supervision, lowering one-step SR from 82.12\% to 75.53\% and increasing the adaptive budget to 4.74 steps.
The \textbf{w/o $\boldsymbol{\mathcal{L}_{\mathrm{rec}}}$} variant removes continuation supervision from recycled student states, reducing adaptive SR from 87.96\% to 86.96\% and supporting the contribution of recovery supervision to task success.
\textbf{Clean-only} trains the scheduler without randomized samples. It achieves 87.56\% SR with 4.05 steps under randomized conditions, close to \textbf{Full (Ours)} at 87.80\% with 3.94 steps, supporting scheduler generalization beyond clean-only training conditions.
Finally, \textbf{Lower thr.} and \textbf{Higher thr.} decrease and increase all default threshold percentile levels by five, respectively. They achieve average SRs of 87.60\% and 87.93\% with 3.08 and 5.85 steps. Both remain within 0.36 percentage points of \textbf{Full (Ours)}, indicating limited success-rate sensitivity over the tested threshold range while allowing substantial adjustment of computation.

\newsavebox{\ablationBBox}

\begin{table}[h]
\caption{Ablations on RoboTwin 2.0.
(a) Fixed-step SR (\%); column numbers indicate denoising steps.
(b) Adaptive SR (\%) and mean denoising steps, including
scheduler generalization and threshold sensitivity.}
\label{tab:ablation}

\begin{center}
\normalfont\footnotesize
\setlength{\tabcolsep}{1.5pt}
\renewcommand{\arraystretch}{1.0}

\sbox{\ablationBBox}{%
\begin{tabular}{@{}l*{15}{c}@{}}
\toprule
\multicolumn{16}{@{}l}{(b) Adaptive inference and threshold sensitivity} \\
\midrule
\multirow{2}{*}{Setting}
& Base
& \multicolumn{2}{c}{w/o $\mathcal{L}_{\mathrm{end}}$}
& \multicolumn{2}{c}{w/o $\mathcal{L}_{\mathrm{int}}$}
& \multicolumn{2}{c}{w/o $\mathcal{L}_{\mathrm{rec}}$}
& \multicolumn{2}{c}{Clean-only}
& \multicolumn{2}{c}{Lower thr.}
& \multicolumn{2}{c}{Higher thr.}
& \multicolumn{2}{c}{Full (Ours)} \\
\cmidrule(lr){2-2}
\cmidrule(lr){3-4}\cmidrule(lr){5-6}
\cmidrule(lr){7-8}\cmidrule(lr){9-10}
\cmidrule(lr){11-12}\cmidrule(lr){13-14}
\cmidrule(lr){15-16}
& SR
& SR & Steps
& SR & Steps
& SR & Steps
& SR & Steps
& SR & Steps
& SR & Steps
& SR & Steps \\
\midrule
Clean & \textbf{88.66}
& 87.34 & 4.95
& 79.32 & 5.36
& 87.40 & 3.99
& 87.94 & 3.96
& 87.98 & 2.88
& 88.04 & 5.87
& 88.12 & 4.02 \\
Random & 87.02
& 87.10 & 4.52
& 76.86 & 5.17
& 86.52 & 3.45
& 87.56 & 4.05
& 87.22 & 3.27
& \textbf{87.82} & 5.83
& 87.80 & 3.94 \\
\midrule
Avg. & 87.84
& 87.22 & 4.74
& 78.09 & 5.27
& 86.96 & 3.72
& 87.75 & 4.01
& 87.60 & 3.08
& 87.93 & 5.85
& \textbf{87.96} & 3.98 \\
\bottomrule
\end{tabular}%
}

\begin{adjustbox}{width=0.9\linewidth}
\begin{minipage}{\wd\ablationBBox}
\centering
\normalfont\footnotesize
\setlength{\tabcolsep}{1.5pt}
\renewcommand{\arraystretch}{1.0}

\begin{tabular*}{\linewidth}
{@{\extracolsep{\fill}}l*{13}{c}@{}}
\toprule
\multicolumn{14}{@{}l}{(a) Fixed-step evaluation} \\
\midrule
\multirow{2}{*}{Setting}
& Base
& \multicolumn{3}{c}{w/o $\mathcal{L}_{\mathrm{end}}$}
& \multicolumn{3}{c}{w/o $\mathcal{L}_{\mathrm{int}}$}
& \multicolumn{3}{c}{w/o $\mathcal{L}_{\mathrm{rec}}$}
& \multicolumn{3}{c}{Full (Ours)} \\
\cmidrule(lr){2-2}\cmidrule(lr){3-5}
\cmidrule(lr){6-8}\cmidrule(lr){9-11}
\cmidrule(lr){12-14}
& 10
& 1 & 2 & 10
& 1 & 2 & 10
& 1 & 2 & 10
& 1 & 2 & 10 \\
\midrule
Clean & 88.66
& 76.70 & 86.94 & 87.40
& 82.18 & 79.66 & 80.20
& 84.20 & 86.30 & 87.96
& 84.14 & 87.50 & \textbf{88.78} \\
Random & 87.02
& 74.36 & 84.24 & 87.16
& 79.26 & 78.22 & 79.24
& 80.52 & 86.12 & \textbf{87.40}
& 80.10 & 84.92 & \textbf{87.40} \\
\midrule
Avg. & 87.84
& 75.53 & 85.59 & 87.28
& 80.72 & 78.94 & 79.72
& 82.36 & 86.21 & 87.68
& 82.12 & 86.21 & \textbf{88.09} \\
\bottomrule
\end{tabular*}

\par\smallskip

\noindent\usebox{\ablationBBox}

\end{minipage}
\end{adjustbox}
\end{center}
\end{table}

\section{Conclusion}
\label{sec:conclusion}

We presented AnyStep WAM for tunable-budget generation and adaptive computation allocation. Budget-aligned integral flow-map distillation learns finite-time transitions from frozen-teacher trajectories using shared LoRA adapters, without estimating student temporal derivatives. A lightweight scheduler selects the smallest predicted-feasible budget from difficulty and budget-dependent fidelity estimates, while preview recycling reuses the preview computation. Across Motus, FastWAM, and LingBotVA, AnyStep improves one-step success rates. Adaptive inference reduces denoising steps and latency on RoboTwin 2.0 and six real-world tasks while maintaining comparable or higher average success rates. These results support coupling tunable-budget generation with fidelity-aware scheduling for efficient WAM inference.

\subsection*{AI use statement}

We used generative AI tools to assist with language polishing, including improving grammar, clarity, and readability. All AI-assisted edits were reviewed and revised by the authors. We take full responsibility for the final content of this paper, including its claims, results, and conclusions.

\bibliography{iclr2027_conference}
\bibliographystyle{iclr2027_conference}

\newpage
\appendix

\section{World Action Model Formulations}
\label{app:wam_formulation}
A visuomotor policy models
$p(\va_{1:H}\mid\vo,\vl)$.
For WAMs that explicitly generate future visual states
$\vz_{1:T}$ and actions, this policy can be expressed as
the marginal of their joint distribution:
\begin{equation}
    p(\va_{1:H}\mid\vo,\vl)
    =
    \int
    p(\va_{1:H},\vz_{1:T}\mid\vo,\vl)
    \,\mathrm{d}\vz_{1:T}.
    \label{eq:wam-factorization}
\end{equation}
This identity does not prescribe a generation order.
Video and actions may be generated
jointly~\citep{bi2026motus} or through interleaved
autoregressive prediction~\citep{li2026causal}.

Alternatively, future visual prediction can serve as an auxiliary
training objective, supporting action generation without explicit
future-video denoising at inference~\citep{yuan2026fast}.
Thus, predictive visual modeling during training does not
necessarily imply video generation during deployment.
Our framework accommodates these distinctions: adaptation and
budget allocation apply to both video and action denoising when
both are used at inference, and to action denoising alone otherwise.

\section{Flow Matching and Flow Maps}
\label{app:appendix_prelim_flow}

\textbf{Flow matching.}
Flow matching (FM) defines a continuous transport between noise and data distributions~\citep{lipman2023flowmatchinggenerativemodeling}.
Let $\vx_t$ denote the modeled state at $t\in[0,1]$, where $\vx_1$ is noise and $x_0$ is a data sample. Conditioned on context $\vc$, the probability-flow ODE is
\begin{equation}
    \frac{d \vx_t}{dt}=\vv(\vx_t,t;\vc),
    \label{eq:pf_ode}
\end{equation}
where $\vv$ is the instantaneous velocity field. Standard generation solves Eq.~\ref{eq:pf_ode} from $t=1$ to $t=0$ using multiple network evaluations.

\textbf{Flow maps and average velocity.}
A flow map instead represents a finite-time transition along the same ODE trajectory. Let $\Phi_{r\leftarrow t}(\vx_t)=\vx_r$ for $1\geq t\geq r\geq0$. It satisfies
$\Phi_{t\leftarrow t}(x)=x$ and
$\Phi_{q\leftarrow r}\circ\Phi_{r\leftarrow t}
=\Phi_{q\leftarrow t}$.
The corresponding average velocity over $[r,t]$ is
\begin{equation}
        \bar{\vv}(\vx_t,t,r;\vc) = \frac{1}{t-r} \int_r^t \vv(\vx_s,s;\vc)\,ds,
    \label{eq:appendix_avg_velocity}
\end{equation}
such that $\Phi_{r\leftarrow t}(\vx_t)=\vx_t-(t-r)\bar{\vv}(\vx_t,t,r;\vc)$.
A neural flow-map model therefore predicts an average velocity
$u_\theta$ and parameterizes
\begin{equation}
    f_\vtheta(\vx_t,t,r;\vc)
    =
    \vx_t-(t-r)\vu_\vtheta(\vx_t,t,r;\vc).
    \label{eq:appendix_flow_map}
\end{equation}
Conditioning on both endpoints $(t,r)$ allows one model to support different inference budgets.

\textbf{Derivative-based flow-map supervision.}
MeanFlow~\citep{geng2026mean} derives a local identity relating
average and instantaneous velocities,
\begin{equation}
    \vu_{\mathrm{tgt}}=\vv(\vx_t,t;\vc)-(t-r)\frac{d}{dt}\vu_\vtheta(\vx_t,t,r;\vc),
    \label{eq:appendix_meanflow_identity}
\end{equation}
and trains
\begin{equation}
    \mathcal{L}_{\mathrm{MF}}
    =\mathbb{E}\left[\left\|\vu_\vtheta-\operatorname{sg}(\vu_{\mathrm{tgt}})\right\|_2^2\right].
    \label{eq:appendix_meanflow_loss}
\end{equation}
The total derivative in Eq.~\ref{eq:appendix_meanflow_identity} is computed through a Jacobian-vector product (JVP). AnyFlow~\citep{gu2026anyflow} instead approximates it using a central finite difference to avoid explicit JVP computation. Both formulations construct the flow-map supervision through local derivative information of the learned average-velocity field. 
Such derivative-dependent objectives can be sensitive to optimization and prediction noise~\citep{geng2026improvedmeanflowschallenges, Tu_2026_CVPR, gu2026anyflow}. 

\section{Implementation Details}
\label{appendix:implementation}

\paragraph{RoboTwin simulation.}
We use the official RoboTwin 2.0 demonstration dataset~\citep{chen2025robotwin}. For each of the 50 tasks, the dataset contains 50 demonstration episodes under the clean setting and 500 under the randomized setting. We instantiate AnyStep on three publicly available backbones: Motus~\citep{bi2026motus}, FastWAM~\citep{yuan2026fast}, and LingBotVA~\citep{li2026causal}. For each backbone, we train a single multi-task AnyStep model shared across all 50 tasks. Clean and randomized demonstrations are sampled with equal probability during distillation.

\paragraph{AnyStep adaptation.}
During AnyStep distillation, all original backbone and vision-language model parameters remain frozen, and only the introduced LoRA adapters are optimized. The same adapters are shared across tasks and denoising budgets. We set the LoRA rank to 128 for all adapted layers across the three backbones. The training objective combines instantaneous-velocity, endpoint, budget-aligned interval, and recycled-state recovery supervision,
with weights $\lambda_{\mathrm{FM}}=0.20$,
$\lambda_{\mathrm{end}}=0.50$,
$\lambda_{\mathrm{int}}=0.30$, and
$\lambda_{\mathrm{rec}}=0.20$, respectively.
We use these weights for all three backbones.
Table~\ref{tab:anystep_training_hyperparameters} summarizes the
distillation hyperparameters and trainable parameter counts.

\begin{table}[h]
    \centering
    \caption{Hyperparameters and trainable parameter counts for
    LoRA-based AnyStep distillation. All original model parameters
    remain frozen during this stage.}
    \label{tab:anystep_training_hyperparameters}
    \vspace{0.15in}
    \begin{tabular}{lc}
        \toprule
        Configuration & Value \\
        \midrule
        Optimizer & AdamW \\
        Learning rate & $2\times10^{-5}$ \\
        Weight decay & $0.01$ \\
        Learning-rate schedule & Linear schedule \\
        Training iterations & $8{,}000$ \\
        Training GPUs & $4\times$ NVIDIA H100 \\
        Batch size per GPU & $8$ \\
        Global batch size & $32$ \\
        \midrule
        LoRA rank & $128$ \\
        $\lambda_{\mathrm{FM}}$ & $0.20$ \\
        $\lambda_{\mathrm{end}}$ & $0.50$ \\
        $\lambda_{\mathrm{int}}$ & $0.30$ \\
        $\lambda_{\mathrm{rec}}$ & $0.20$ \\
        \midrule
        Trainable LoRA parameters (Motus) & $286.39$M \\
        Trainable LoRA parameters (FastWAM) & $104.40$M \\
        Trainable LoRA parameters (LingBotVA) & $232.91$M \\
        \bottomrule
    \end{tabular}
\end{table}

\begin{table}[h]
    \centering
    \caption{Training hyperparameters of the adaptive scheduler.
    The teacher and AnyStep student remain frozen during
    scheduler training.}
    \label{tab:scheduler_hyperparameters}
    \vspace{0.15in}
    \begin{tabular}{lc}
        \toprule
        Configuration & Value \\
        \midrule
        Optimizer & AdamW \\
        Learning rate & $1\times10^{-4}$ \\
        Weight decay & $0.01$ \\
        Learning-rate schedule & Cosine decay \\
        Number of training GPUs & $4$ \\
        Batch size per GPU & $64$ \\
        Global batch size & $256$ \\
        Training iterations & $4{,}000$ \\
        Prediction loss & Smooth L1 \\
        Risk prediction loss weight & $1.0$ \\
        Fidelity prediction loss weight & $1.0$ \\
        \bottomrule
    \end{tabular}
\end{table}

\paragraph{Real-world experiments.}
Our real-world dataset covers six manipulation tasks, with 50
demonstrations collected per task, yielding 300 demonstrations in total.
For each backbone, we first train a shared multi-task base model using
demonstrations from all six tasks. We then freeze the resulting model
and optimize only the newly introduced LoRA adapters during AnyStep
adaptation. The base model and its AnyStep variant use the same
demonstration dataset.

\paragraph{Candidate budgets and backbone-specific execution.}
We use the candidate budget set $\mathcal{K}=\{1,2,\ldots,10\}$ for
all three backbones.
Motus denoises video and actions jointly, with both branches following
a synchronized $K$-step schedule.
FastWAM omits future-video denoising at inference, so adaptive budget
selection and preview recycling apply only to its action branch.
LingBotVA follows a sequential video-to-action procedure, which first denoises video and then
generates the action conditioned on the video output.

\begin{figure}[h]
    \centering
    \includegraphics[width=0.65\linewidth, trim={5mm 5mm 5mm 5mm}, clip]{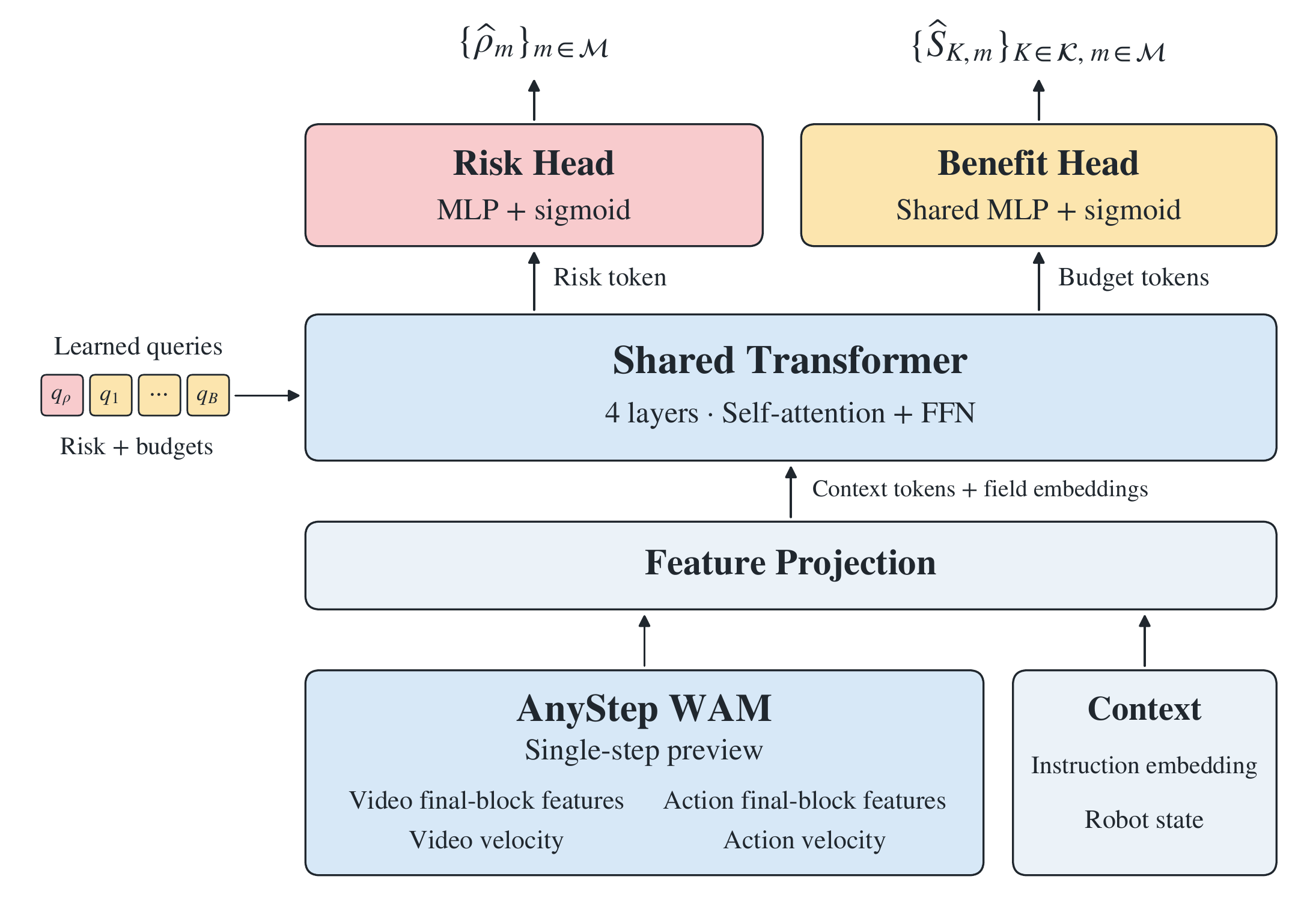}
    \caption{Internal architecture of the preview-based adaptive scheduler. Features and velocity predictions from a single-step AnyStep WAM preview are fused with instruction and robot-state context, projected into context tokens, and processed jointly with learned risk and budget queries by a shared Transformer. Query-specific heads predict modality-wise execution risk and the expected student-teacher similarity for each candidate denoising budget.}
    \label{fig:appendix_preview_scheduler}
\end{figure}

\paragraph{Adaptive scheduler data and training.}
After AnyStep distillation, we freeze the student and construct the
scheduler training set offline using the same training demonstrations.
For each sampled context and initial noise, we extract features from
the one-step preview and perform offline teacher-student evaluations
to obtain teacher-derived risk labels and budget-dependent fidelity
labels. Candidate budgets are evaluated using the preview-recycled
execution procedure described in Section~\ref{sec:adaptive_inference}.
This process requires no additional environment interaction or
demonstration collection.
Both the teacher and student remain frozen while the scheduler is
trained to predict the offline labels using equally weighted
Smooth L1 losses for risk and fidelity prediction.
The scheduler architecture is described in Figure~\ref{fig:appendix_preview_scheduler}, and its training
hyperparameters are listed in Table~\ref{tab:scheduler_hyperparameters}.

As shown in Figure~\ref{fig:appendix_preview_scheduler}, the scheduler predicts these offline targets from features available
in a single one-step preview.
For joint video-action models, its inputs comprise six feature
groups: video/action final-block feature summaries, video/action
predicted-velocity summaries, normalized robot state, a pooled
instruction embedding.
A four-layer Transformer processes the projected context tokens
together with one risk query and $B=|\mathcal K|$ budget queries.
The risk head predicts $\{\widehat\rho_m\}_{m\in\mathcal M}$,
while the benefit head predicts a
$B\times|\mathcal M|$ matrix
$\{\widehat S_{K,m}\}_{K\in\mathcal K,m\in\mathcal M}$.
Both heads use sigmoid outputs and are trained with Smooth L1
losses against their respective offline targets.
This allows a single preview to estimate difficulty and compare
candidate budgets without online teacher evaluations or
candidate rollouts.

\section{Training-free gating baseline}
\label{appendix:Training-free gating baseline}
We first execute complete one-step action denoising as a probe. For each arm,
we compute the relative contribution of each joint to the predicted motion,
\begin{equation}
    p_{t,j}
    =
    \frac{|q_{t+1,j}-q_{t,j}|}
    {\sum_{k=1}^{6}|q_{t+1,k}-q_{t,k}|+\epsilon},
\end{equation}
and define the action-coordination metric as
\begin{equation}
    C_{\mathrm{raw}}
    =
    \max_{\mathrm{arm}\in\{L,R\}}
    Q_{0.9}
    \left(
        \sum_{j=1}^{6}
        |p_{t+1,j}-p_{t,j}|
    \right).
\end{equation}
Thus, $C_{\mathrm{raw}}$ measures how strongly the dominant moving joints
change within the one-step action proposal.

For the visual metric, we compute the patch-wise RGB difference
$D(\mathbf{I}_a,\mathbf{I}_b)$ between two real observations as the $90$-th
percentile of the mean absolute differences over a $4\times4$ patch grid.
Using the latest three observations, the visual irregularity is
\begin{equation}
    V_{\mathrm{raw}}
    =
    \frac{
        |D(\mathbf{I}_{t-1},\mathbf{I}_{t})
        -D(\mathbf{I}_{t-2},\mathbf{I}_{t-1})|
    }{
        \frac{1}{2}
        \left[
        D(\mathbf{I}_{t-1},\mathbf{I}_{t})
        +D(\mathbf{I}_{t-2},\mathbf{I}_{t-1})
        \right]+\epsilon
    }.
\end{equation}
The two metrics are normalized using frozen ECDFs:
\begin{equation}
    C=F_C(C_{\mathrm{raw}}),
    \qquad
    V=1-F_V(V_{\mathrm{raw}}),
    \qquad
    S_{CV}=\frac{1}{2}C+\frac{1}{2}V.
\end{equation}
Finally, with $u=F_{CV}(S_{CV})$, the denoising budget is selected as
\begin{equation}
    N
    =
    \min\left(
        K,\,
        \left\lfloor K u^\gamma\right\rfloor+1
    \right),
\end{equation}
where $K=10$ and $\gamma=1.6965$. If $N=1$, the probe is directly reused;
otherwise, a complete $N$-step trajectory is rerun from the same initial
noise.

\section{Fixed denoise step results on RoboTwin}
\label{appendix:fixed_denoise_step_results}

\begin{table}[h]
\begin{center}
\caption{Complete fixed-step SR (\%) on RoboTwin 2.0. Base denotes the original models; AnyFlow and AnyStep denote the corresponding training methods. Avg. averages the clean and randomized settings.}
\label{tab:appendix_robotwin_fixedstep_full}
\vspace{6pt}
\normalfont
\setlength{\tabcolsep}{3pt}
\renewcommand{\arraystretch}{1.0}
\begin{adjustbox}{max width=\linewidth}
\begin{tabular}{@{}cl*{9}{r}@{}}
\toprule
\multirow{2}{*}{Steps}
& \multirow{2}{*}{Method}
& \multicolumn{3}{c}{Motus}
& \multicolumn{3}{c}{FastWAM}
& \multicolumn{3}{c}{LingBotVA} \\
\cmidrule(lr){3-5}\cmidrule(lr){6-8}\cmidrule(lr){9-11}
& & Clean & Random & Avg.
  & Clean & Random & Avg.
  & Clean & Random & Avg. \\
\midrule
\multirow{3}{*}{1}
& Base
& 76.00 & 74.10 & 75.05 & 77.26 & 77.14 & 77.20 & 74.36 & 73.18 & 73.77 \\
& AnyFlow
& 13.20 & 14.30 & 13.75 & 32.60 & 31.42 & 32.01 & 51.46 & 48.46 & 49.96 \\
& AnyStep (Ours)
& \textbf{84.14} & \textbf{80.10} & \textbf{82.12} & \textbf{89.12} & \textbf{89.44} & \textbf{89.28} & \textbf{83.80} & \textbf{81.62} & \textbf{82.71} \\
\midrule
\multirow{3}{*}{2}
& Base
& 82.60 & 83.92 & 83.26 & 84.52 & 83.70 & 84.11 & 85.38 & 85.26 & 85.32 \\
& AnyFlow
& 52.92 & 51.80 & 52.36 & 38.24 & 36.68 & 37.46 & 83.28 & 79.26 & 81.27 \\
& AnyStep (Ours)
& \textbf{87.50} & \textbf{84.92} & \textbf{86.21} & \textbf{90.20} & \textbf{89.32} & \textbf{89.76} & \textbf{88.64} & \textbf{89.28} & \textbf{88.96} \\
\midrule
\multirow{3}{*}{4}
& Base
& 84.20 & 83.60 & 83.90 & 87.80 & 86.28 & 87.04 & 86.92 & 85.38 & 86.15 \\
& AnyFlow
& 81.34 & 81.80 & 81.57 & 44.44 & 40.66 & 42.55 & 85.24 & 83.32 & 84.28 \\
& AnyStep (Ours)
& \textbf{87.40} & \textbf{86.68} & \textbf{87.04} & \textbf{90.72} & \textbf{90.40} & \textbf{90.56} & \textbf{89.88} & \textbf{89.26} & \textbf{89.57} \\
\midrule
\multirow{3}{*}{6}
& Base
& 86.60 & 85.20 & 85.90 & 88.14 & 87.34 & 87.74 & 88.48 & 88.12 & 88.30 \\
& AnyFlow
& 84.78 & 84.92 & 84.85 & 61.68 & 60.92 & 61.30 & 90.02 & 86.24 & 88.13 \\
& AnyStep (Ours)
& \textbf{87.56} & \textbf{86.32} & \textbf{86.94} & \textbf{91.60} & \textbf{90.52} & \textbf{91.06} & \textbf{90.60} & \textbf{90.14} & \textbf{90.37} \\
\midrule
\multirow{3}{*}{8}
& Base
& 85.60 & 85.60 & 85.60 & 90.16 & 89.62 & 89.89 & 90.62 & 89.86 & 90.24 \\
& AnyFlow
& 87.44 & 86.30 & 86.87 & 62.68 & 64.64 & 63.66 & 90.26 & 87.20 & 88.73 \\
& AnyStep (Ours)
& \textbf{87.94} & \textbf{87.34} & \textbf{87.64} & \textbf{92.24} & \textbf{91.76} & \textbf{92.00} & \textbf{91.84} & \textbf{90.72} & \textbf{91.28} \\
\midrule
\multirow{3}{*}{10}
& Base
& 88.66 & 87.02 & 87.84 & 91.88 & \textbf{91.78} & 91.83 & 90.42 & 90.36 & 90.39 \\
& AnyFlow
& 87.62 & 87.14 & 87.38 & 78.42 & 75.60 & 77.01 & 90.12 & 89.28 & 89.70 \\
& AnyStep (Ours)
& \textbf{88.78} & \textbf{87.40} & \textbf{88.09} & \textbf{92.48} & 91.40 & \textbf{91.94} & \textbf{92.92} & \textbf{91.36} & \textbf{92.14} \\
\bottomrule
\end{tabular}
\end{adjustbox}
\end{center}
\end{table}

Table~\ref{tab:appendix_robotwin_fixedstep_full} reports additional results on RoboTwin 2.0 across fixed denoising budgets. AnyFlow exhibits a pronounced performance drop at small budgets. Action generation requires precise predictions, as small errors during grasping or contact can lead to task failure. The training instability observed in our experiments may compromise the accuracy of large denoising updates, making this issue particularly severe under few-step inference. Increasing the budget enables smaller updates and further refinement, which partially mitigate these errors, consistent with AnyFlow's improving success rates as the number of steps increases.

\section{Additional Ablations on FastWAM and LingBotVA}
\label{app:additional_ablation_backbones}

We extend the Motus ablations in Table~\ref{tab:ablation} to FastWAM and LingBotVA using the same variant definitions. Avg. denotes the arithmetic mean of clean and randomized results. Lower thr. and Higher thr. shift all applicable default percentile thresholds by $-5$ and $+5$ percentile points, respectively. For FastWAM, these changes apply only to the action thresholds.

\newsavebox{\fastwamAblationBBox}
\begin{table}[h]
\caption{Ablations on FastWAM using RoboTwin 2.0. (a) Fixed-step SR (\%); column numbers indicate denoising steps. (b) Adaptive SR (\%) and mean denoising steps. Bold marks the highest SR in each row within each panel.}
\label{tab:ablation_fastwam}
\vspace{0.15in}
\centering
\normalfont\footnotesize
\setlength{\tabcolsep}{1.5pt}
\renewcommand{\arraystretch}{1.0}
\sbox{\fastwamAblationBBox}{%
\begin{tabular}{@{}l*{15}{c}@{}}
\toprule
\multicolumn{16}{@{}l}{(b) Adaptive inference and threshold sensitivity} \\
\midrule
\multirow{2}{*}{Setting}
& Base
& \multicolumn{2}{c}{w/o $\mathcal{L}_{\mathrm{end}}$}
& \multicolumn{2}{c}{w/o $\mathcal{L}_{\mathrm{int}}$}
& \multicolumn{2}{c}{w/o $\mathcal{L}_{\mathrm{rec}}$}
& \multicolumn{2}{c}{Clean-only}
& \multicolumn{2}{c}{Lower thr.}
& \multicolumn{2}{c}{Higher thr.}
& \multicolumn{2}{c}{Full (Ours)} \\
\cmidrule(lr){2-2}
\cmidrule(lr){3-4}\cmidrule(lr){5-6}
\cmidrule(lr){7-8}\cmidrule(lr){9-10}
\cmidrule(lr){11-12}\cmidrule(lr){13-14}
\cmidrule(lr){15-16}
& SR & SR & Steps & SR & Steps & SR & Steps
& SR & Steps & SR & Steps & SR & Steps & SR & Steps \\
\midrule
Clean & 91.88 & 91.54 & 5.21 & 82.54 & 6.43 & 91.86 & 4.67 & 91.94 & 4.99 & 92.08 & 3.92 & \textbf{92.14} & 6.11 & 92.12 & 5.02 \\
Random & \textbf{91.78} & 90.48 & 5.92 & 80.72 & 6.12 & 90.58 & 4.92 & 90.98 & 5.27 & 90.84 & 3.97 & 91.58 & 5.98 & 91.06 & 5.03 \\
\midrule
Avg. & 91.83 & 91.01 & 5.57 & 81.63 & 6.28 & 91.22 & 4.80 & 91.46 & 5.13 & 91.46 & 3.95 & \textbf{91.86} & 6.05 & 91.59 & 5.03 \\
\bottomrule
\end{tabular}%
}
\begin{adjustbox}{width=0.9\linewidth}
\begin{minipage}{\wd\fastwamAblationBBox}
\centering
\normalfont\footnotesize
\setlength{\tabcolsep}{1.5pt}
\renewcommand{\arraystretch}{1.0}
\begin{tabular*}{\linewidth}{@{\extracolsep{\fill}}l*{13}{c}@{}}
\toprule
\multicolumn{14}{@{}l}{(a) Fixed-step evaluation} \\
\midrule
\multirow{2}{*}{Setting}
& Base
& \multicolumn{3}{c}{w/o $\mathcal{L}_{\mathrm{end}}$}
& \multicolumn{3}{c}{w/o $\mathcal{L}_{\mathrm{int}}$}
& \multicolumn{3}{c}{w/o $\mathcal{L}_{\mathrm{rec}}$}
& \multicolumn{3}{c}{Full (Ours)} \\
\cmidrule(lr){2-2}\cmidrule(lr){3-5}
\cmidrule(lr){6-8}\cmidrule(lr){9-11}\cmidrule(lr){12-14}
& 10 & 1 & 2 & 10 & 1 & 2 & 10 & 1 & 2 & 10 & 1 & 2 & 10 \\
\midrule
Clean & 91.88 & 79.22 & 89.92 & 91.64 & 85.48 & 82.26 & 82.98 & 89.40 & 91.00 & 92.22 & 89.12 & 90.20 & \textbf{92.48} \\
Random & \textbf{91.78} & 76.34 & 89.04 & 91.26 & 84.32 & 80.44 & 81.04 & 89.28 & 89.96 & 91.54 & 89.44 & 89.32 & 91.40 \\
\midrule
Avg. & 91.83 & 77.78 & 89.48 & 91.45 & 84.90 & 81.35 & 82.01 & 89.34 & 90.48 & 91.88 & 89.28 & 89.76 & \textbf{91.94} \\
\bottomrule
\end{tabular*}
\par\smallskip
\noindent\usebox{\fastwamAblationBBox}
\end{minipage}
\end{adjustbox}
\end{table}

\newsavebox{\lingbotvaAblationBBox}
\begin{table}[h]
\caption{Ablations on LingBotVA using RoboTwin 2.0. (a) Fixed-step SR (\%); column numbers indicate denoising steps. (b) Adaptive SR (\%) and mean denoising steps. Bold marks the highest SR in each row within each panel. Base uses 25/50 video/action steps; AnyStep budgets apply to each branch.}
\label{tab:ablation_lingbotva}
\vspace{0.15in}
\centering
\normalfont\footnotesize
\setlength{\tabcolsep}{1.5pt}
\renewcommand{\arraystretch}{1.0}
\sbox{\lingbotvaAblationBBox}{%
\begin{tabular}{@{}l*{15}{c}@{}}
\toprule
\multicolumn{16}{@{}l}{(b) Adaptive inference and threshold sensitivity} \\
\midrule
\multirow{2}{*}{Setting}
& Base
& \multicolumn{2}{c}{w/o $\mathcal{L}_{\mathrm{end}}$}
& \multicolumn{2}{c}{w/o $\mathcal{L}_{\mathrm{int}}$}
& \multicolumn{2}{c}{w/o $\mathcal{L}_{\mathrm{rec}}$}
& \multicolumn{2}{c}{Clean-only}
& \multicolumn{2}{c}{Lower thr.}
& \multicolumn{2}{c}{Higher thr.}
& \multicolumn{2}{c}{Full (Ours)} \\
\cmidrule(lr){2-2}
\cmidrule(lr){3-4}\cmidrule(lr){5-6}
\cmidrule(lr){7-8}\cmidrule(lr){9-10}
\cmidrule(lr){11-12}\cmidrule(lr){13-14}
\cmidrule(lr){15-16}
& SR & SR & Steps & SR & Steps & SR & Steps
& SR & Steps & SR & Steps & SR & Steps & SR & Steps \\
\midrule
Clean & 92.93 & 91.32 & 4.59 & 80.76 & 5.14 & 92.08 & 3.76 & 92.18 & 3.37 & 91.98 & 2.98 & \textbf{92.98} & 5.79 & 92.74 & 3.54 \\
Random & 91.55 & 91.06 & 4.81 & 78.08 & 5.99 & 91.24 & 4.02 & 91.26 & 3.96 & 90.56 & 3.11 & \textbf{91.82} & 6.23 & 91.70 & 3.82 \\
\midrule
Avg. & 92.24 & 91.19 & 4.70 & 79.42 & 5.57 & 91.66 & 3.89 & 91.72 & 3.67 & 91.27 & 3.05 & \textbf{92.40} & 6.01 & 92.22 & 3.68 \\
\bottomrule
\end{tabular}%
}
\begin{adjustbox}{width=0.9\linewidth}
\begin{minipage}{\wd\lingbotvaAblationBBox}
\centering
\normalfont\footnotesize
\setlength{\tabcolsep}{1.5pt}
\renewcommand{\arraystretch}{1.0}
\begin{tabular*}{\linewidth}{@{\extracolsep{\fill}}l*{13}{c}@{}}
\toprule
\multicolumn{14}{@{}l}{(a) Fixed-step evaluation} \\
\midrule
\multirow{2}{*}{Setting}
& Base
& \multicolumn{3}{c}{w/o $\mathcal{L}_{\mathrm{end}}$}
& \multicolumn{3}{c}{w/o $\mathcal{L}_{\mathrm{int}}$}
& \multicolumn{3}{c}{w/o $\mathcal{L}_{\mathrm{rec}}$}
& \multicolumn{3}{c}{Full (Ours)} \\
\cmidrule(lr){2-2}\cmidrule(lr){3-5}
\cmidrule(lr){6-8}\cmidrule(lr){9-11}\cmidrule(lr){12-14}
& 25/50 & 1 & 2 & 10 & 1 & 2 & 10 & 1 & 2 & 10 & 1 & 2 & 10 \\
\midrule
Clean & \textbf{92.93} & 75.24 & 87.94 & 91.58 & 80.28 & 76.62 & 78.84 & 82.16 & 87.32 & 91.86 & 83.80 & 88.64 & 92.92 \\
Random & \textbf{91.55} & 74.02 & 88.20 & 91.24 & 77.16 & 73.14 & 76.28 & 80.04 & 88.84 & 90.82 & 81.62 & 89.28 & 91.36 \\
\midrule
Avg. & \textbf{92.24} & 74.63 & 88.07 & 91.41 & 78.72 & 74.88 & 77.56 & 81.10 & 88.08 & 91.34 & 82.71 & 88.96 & 92.14 \\
\bottomrule
\end{tabular*}
\par\smallskip
\noindent\usebox{\lingbotvaAblationBBox}
\end{minipage}
\end{adjustbox}
\end{table}

\section{RoboTwin Detailed Results}
\label{appendix:all_tasks_results}

Tables~\ref{tab:motus_all_tasks},~\ref{tab:fastwam_all_tasks}, and~\ref{tab:lingbotva_all_tasks} present task-wise comparisons between the base models and our AnyStep variants with adaptive denoising. We report success rates under clean and randomized settings, together with the average denoising steps used by AnyStep. Following their original inference configurations, the base models use their full fixed denoising schedules: 10 steps for Motus~\citep{bi2026motus}, 10 steps for FastWAM~\citep{yuan2026fast}, and 25 and 50 steps for the video and action DiTs of LingBotVA~\citep{li2026causal}, respectively. Our AnyStep variants adaptively select the denoising budget
for each prediction chunk.

\begin{table}[p]
\centering
\caption{\textbf{Task-wise comparison of Motus and AnyStep-Motus across 50 tasks.} We report success rates (SR, \%) under clean and randomized (Rand.) settings, and average denoising steps (Steps) for AnyStep-Motus. The final row reports the unweighted mean across tasks. Bold indicates the highest SR within each row and setting, including ties.}
\label{tab:motus_all_tasks}
\vspace{\baselineskip}
\setlength{\tabcolsep}{4pt}
\renewcommand{\arraystretch}{0.94}
\begin{tabular}{@{}lcccccc@{}}
\toprule
Task & \multicolumn{2}{c}{Motus~\citep{bi2026motus}} & \multicolumn{4}{c}{AnyStep-Motus} \\
\cmidrule(lr){2-3}\cmidrule(lr){4-7}
& Clean & Rand. & \multicolumn{2}{c}{Clean} & \multicolumn{2}{c}{Rand.} \\
\cmidrule(lr){4-5}\cmidrule(lr){6-7}
& SR $\uparrow$ & SR $\uparrow$ & SR $\uparrow$ & Steps $\downarrow$ & SR $\uparrow$ & Steps $\downarrow$ \\
\midrule
Adjust Bottle & \textbf{89} & 93 & \textbf{89} & 5.52 & \textbf{95} & 4.53 \\
Beat Block Hammer & 95 & 88 & \textbf{97} & 3.49 & \textbf{93} & 4.03 \\
Blocks Ranking RGB & 99 & \textbf{97} & \textbf{100} & 2.82 & 95 & 3.11 \\
Blocks Ranking Size & \textbf{75} & 63 & 74 & 3.83 & \textbf{71} & 3.79 \\
Click Alarmclock & \textbf{100} & \textbf{100} & \textbf{100} & 2.55 & \textbf{100} & 2.49 \\
Click Bell & \textbf{100} & \textbf{100} & \textbf{100} & 2.23 & \textbf{100} & 1.78 \\
Dump Bin Bigbin & \textbf{95} & \textbf{91} & \textbf{95} & 3.86 & 90 & 4.30 \\
Grab Roller & \textbf{100} & \textbf{100} & \textbf{100} & 5.39 & \textbf{100} & 5.40 \\
Handover Block & \textbf{86} & \textbf{73} & 78 & 5.87 & 69 & 4.97 \\
Handover Mic & 78 & 63 & \textbf{89} & 3.56 & \textbf{75} & 4.70 \\
Hanging Mug & \textbf{38} & 38 & 34 & 6.46 & \textbf{55} & 6.44 \\
Lift Pot & 96 & \textbf{99} & \textbf{98} & 4.58 & \textbf{99} & 4.57 \\
Move Can Pot & 34 & 74 & \textbf{45} & 3.44 & \textbf{75} & 3.47 \\
Move Pillbottle Pad & 93 & \textbf{96} & \textbf{94} & 3.72 & 93 & 3.37 \\
Move Playingcard Away & \textbf{100} & 96 & \textbf{100} & 3.91 & \textbf{98} & 4.24 \\
Move Stapler Pad & 83 & \textbf{85} & \textbf{87} & 3.72 & 83 & 3.52 \\
Open Laptop & 95 & 91 & \textbf{97} & 3.96 & \textbf{95} & 3.72 \\
Open Microwave & 95 & \textbf{91} & \textbf{96} & 3.57 & 85 & 3.37 \\
Pick Diverse Bottles & \textbf{90} & \textbf{91} & \textbf{90} & 4.59 & 81 & 4.87 \\
Pick Dual Bottles & \textbf{96} & \textbf{90} & \textbf{96} & 4.13 & \textbf{90} & 4.40 \\
Place A2B Left & \textbf{88} & 79 & 85 & 3.97 & \textbf{95} & 3.58 \\
Place A2B Right & \textbf{91} & 87 & 82 & 3.59 & \textbf{95} & 3.50 \\
Place Bread Basket & 91 & 94 & \textbf{94} & 3.96 & \textbf{95} & 3.60 \\
Place Bread Skillet & 86 & 83 & \textbf{87} & 5.27 & \textbf{88} & 4.89 \\
Place Burger Fries & \textbf{98} & \textbf{98} & 95 & 2.71 & \textbf{98} & 2.80 \\
Place Can Basket & \textbf{81} & \textbf{76} & 78 & 3.56 & 69 & 3.84 \\
Place Cans Plasticbox & \textbf{98} & \textbf{94} & 91 & 3.69 & 88 & 4.36 \\
Place Container Plate & \textbf{98} & 99 & \textbf{98} & 2.86 & \textbf{100} & 2.70 \\
Place Dual Shoes & \textbf{93} & \textbf{87} & 92 & 4.85 & \textbf{87} & 5.57 \\
Place Empty Cup & \textbf{99} & 98 & \textbf{99} & 2.94 & \textbf{99} & 2.54 \\
Place Fan & 91 & \textbf{87} & \textbf{94} & 4.02 & 85 & 4.05 \\
Place Mouse Pad & \textbf{66} & 68 & 58 & 3.98 & \textbf{75} & 3.70 \\
Place Object Basket & 81 & \textbf{87} & \textbf{84} & 4.78 & 86 & 4.54 \\
Place Object Scale & \textbf{88} & 85 & 81 & 5.14 & \textbf{89} & 4.30 \\
Place Object Stand & \textbf{98} & \textbf{97} & \textbf{98} & 3.06 & 95 & 3.34 \\
Place Phone Stand & 87 & 86 & \textbf{91} & 4.27 & \textbf{95} & 3.63 \\
Place Shoe & 99 & 97 & \textbf{100} & 3.07 & \textbf{100} & 2.87 \\
Press Stapler & 93 & \textbf{98} & \textbf{96} & 3.70 & \textbf{98} & 2.72 \\
Put Bottles Dustbin & 81 & 79 & \textbf{83} & 5.41 & \textbf{82} & 5.18 \\
Put Object Cabinet & \textbf{88} & \textbf{71} & 60 & 6.31 & 55 & 5.67 \\
Rotate Qrcode & \textbf{89} & \textbf{73} & 83 & 5.37 & 66 & 5.18 \\
Scan Object & 67 & \textbf{66} & \textbf{73} & 5.87 & 64 & 5.75 \\
Shake Bottle & \textbf{100} & 97 & \textbf{100} & 3.85 & \textbf{99} & 3.60 \\
Shake Bottle Horizontally & \textbf{100} & \textbf{98} & \textbf{100} & 3.78 & 95 & 4.22 \\
Stack Blocks Three & 91 & 95 & \textbf{97} & 2.73 & \textbf{96} & 2.78 \\
Stack Blocks Two & \textbf{100} & \textbf{98} & \textbf{100} & 2.34 & 97 & 2.69 \\
Stack Bowls Three & \textbf{79} & \textbf{87} & 75 & 4.15 & \textbf{87} & 4.32 \\
Stack Bowls Two & 98 & \textbf{98} & \textbf{99} & 2.92 & 97 & 3.45 \\
Stamp Seal & 93 & 92 & \textbf{97} & 2.95 & \textbf{97} & 3.01 \\
Turn Switch & \textbf{84} & \textbf{78} & 77 & 4.72 & 76 & 3.47 \\
\midrule
\textbf{Average} & \textbf{88.66} & 87.02 & 88.12 & 4.02 & \textbf{87.8} & 3.94 \\
\bottomrule
\end{tabular}
\end{table}

\begin{table}[p]
\centering
\caption{\textbf{Task-wise comparison of FastWAM and AnyStep-FastWAM across 50 tasks.} We report success rates (SR, \%) under clean and randomized (Rand.) settings, and average denoising steps (Steps) for AnyStep-FastWAM. The final row reports the aggregate results. Bold indicates the highest SR within each row and setting, including ties.}
\label{tab:fastwam_all_tasks}
\vspace{\baselineskip}
\setlength{\tabcolsep}{4pt}
\renewcommand{\arraystretch}{0.94}
\begin{tabular}{@{}lcccccc@{}}
\toprule
Task & \multicolumn{2}{c}{FastWAM~\citep{yuan2026fast}} & \multicolumn{4}{c}{AnyStep-FastWAM} \\
\cmidrule(lr){2-3}\cmidrule(lr){4-7}
& Clean & Rand. & \multicolumn{2}{c}{Clean} & \multicolumn{2}{c}{Rand.} \\
\cmidrule(lr){4-5}\cmidrule(lr){6-7}
& SR $\uparrow$ & SR $\uparrow$ & SR $\uparrow$ & Steps $\downarrow$ & SR $\uparrow$ & Steps $\downarrow$ \\
\midrule
Adjust Bottle & \textbf{100} & \textbf{100} & \textbf{100} & 5.57 & \textbf{100} & 6.18 \\
Beat Block Hammer & 99 & 97 & \textbf{100} & 4.50 & \textbf{100} & 4.81 \\
Blocks Ranking RGB & \textbf{100} & \textbf{100} & \textbf{100} & 4.00 & 98 & 4.15 \\
Blocks Ranking Size & \textbf{94} & \textbf{98} & 93 & 4.55 & 92 & 4.81 \\
Click Alarmclock & \textbf{100} & \textbf{100} & \textbf{100} & 5.82 & \textbf{100} & 5.53 \\
Click Bell & \textbf{100} & \textbf{100} & \textbf{100} & 5.26 & \textbf{100} & 5.18 \\
Dump Bin Bigbin & 97 & \textbf{96} & \textbf{99} & 4.83 & 95 & 4.72 \\
Grab Roller & \textbf{100} & \textbf{100} & \textbf{100} & 4.31 & \textbf{100} & 4.42 \\
Handover Block & \textbf{95} & \textbf{81} & \textbf{95} & 6.13 & \textbf{81} & 6.25 \\
Handover Mic & 99 & \textbf{100} & \textbf{100} & 6.75 & \textbf{100} & 6.23 \\
Hanging Mug & 58 & \textbf{62} & \textbf{63} & 6.22 & 61 & 6.93 \\
Lift Pot & \textbf{100} & \textbf{100} & \textbf{100} & 4.17 & \textbf{100} & 4.60 \\
Move Can Pot & \textbf{90} & 88 & 86 & 5.43 & \textbf{91} & 5.17 \\
Move Pillbottle Pad & \textbf{100} & 99 & \textbf{100} & 4.69 & \textbf{100} & 4.52 \\
Move Playingcard Away & \textbf{100} & \textbf{100} & \textbf{100} & 5.61 & \textbf{100} & 5.25 \\
Move Stapler Pad & 77 & 64 & \textbf{80} & 5.23 & \textbf{66} & 4.75 \\
Open Laptop & 98 & \textbf{100} & \textbf{99} & 4.51 & 99 & 4.09 \\
Open Microwave & 62 & \textbf{45} & \textbf{79} & 4.39 & 36 & 5.20 \\
Pick Diverse Bottles & 80 & 85 & \textbf{87} & 4.96 & \textbf{87} & 5.17 \\
Pick Dual Bottles & \textbf{100} & 96 & 99 & 4.84 & \textbf{98} & 5.27 \\
Place A2B Left & \textbf{95} & 93 & 93 & 5.39 & \textbf{95} & 5.30 \\
Place A2B Right & 93 & \textbf{99} & \textbf{94} & 5.01 & 95 & 5.23 \\
Place Bread Basket & 91 & \textbf{93} & \textbf{93} & 5.00 & \textbf{93} & 5.43 \\
Place Bread Skillet & 90 & \textbf{93} & \textbf{93} & 6.64 & 91 & 6.64 \\
Place Burger Fries & \textbf{96} & \textbf{99} & 94 & 4.84 & 97 & 5.29 \\
Place Can Basket & \textbf{71} & \textbf{69} & 68 & 5.71 & 65 & 5.19 \\
Place Cans Plasticbox & \textbf{99} & 96 & 98 & 4.75 & \textbf{98} & 4.74 \\
Place Container Plate & 96 & \textbf{100} & \textbf{100} & 4.79 & \textbf{100} & 4.60 \\
Place Dual Shoes & \textbf{94} & 88 & 91 & 5.26 & \textbf{90} & 5.67 \\
Place Empty Cup & \textbf{100} & \textbf{100} & 98 & 3.87 & \textbf{100} & 3.94 \\
Place Fan & 96 & \textbf{96} & \textbf{99} & 4.87 & 92 & 5.13 \\
Place Mouse Pad & 83 & \textbf{89} & \textbf{90} & 4.90 & 88 & 4.96 \\
Place Object Basket & \textbf{89} & \textbf{88} & 82 & 5.40 & 82 & 5.10 \\
Place Object Scale & \textbf{90} & \textbf{97} & 88 & 5.25 & 96 & 5.36 \\
Place Object Stand & \textbf{90} & 94 & 86 & 4.06 & \textbf{95} & 3.99 \\
Place Phone Stand & 97 & \textbf{99} & \textbf{98} & 5.52 & \textbf{99} & 5.59 \\
Place Shoe & 96 & \textbf{99} & \textbf{97} & 4.69 & 98 & 4.45 \\
Press Stapler & 90 & \textbf{97} & \textbf{94} & 4.41 & 96 & 4.00 \\
Put Bottles Dustbin & \textbf{95} & \textbf{90} & 87 & 5.82 & \textbf{90} & 4.87 \\
Put Object Cabinet & \textbf{94} & \textbf{89} & 85 & 5.33 & 87 & 5.07 \\
Rotate Qrcode & \textbf{93} & \textbf{89} & 92 & 5.47 & 88 & 5.49 \\
Scan Object & 89 & \textbf{92} & \textbf{93} & 6.86 & 89 & 6.27 \\
Shake Bottle & \textbf{100} & \textbf{100} & \textbf{100} & 4.78 & \textbf{100} & 4.72 \\
Shake Bottle Horizontally & \textbf{100} & \textbf{100} & \textbf{100} & 4.63 & \textbf{100} & 4.57 \\
Stack Blocks Three & \textbf{95} & \textbf{97} & 90 & 4.16 & \textbf{97} & 4.03 \\
Stack Blocks Two & \textbf{100} & \textbf{100} & \textbf{100} & 3.44 & 98 & 3.77 \\
Stack Bowls Three & \textbf{80} & \textbf{81} & 75 & 5.22 & 77 & 5.04 \\
Stack Bowls Two & 92 & \textbf{98} & \textbf{95} & 4.74 & 93 & 4.50 \\
Stamp Seal & \textbf{90} & \textbf{94} & 89 & 4.75 & 92 & 4.44 \\
Turn Switch & 61 & 59 & \textbf{64} & 3.67 & \textbf{68} & 4.89 \\
\midrule
\textbf{Average} & 91.88 & \textbf{91.78} & \textbf{92.12} & 5.02 & 91.06 & 5.03 \\
\bottomrule
\end{tabular}
\end{table}

\begin{table}[p]
\centering
\caption{\textbf{Task-wise comparison of LingBotVA and AnyStep-LingBotVA across 50 tasks.} We report success rates (SR, \%) under clean and randomized (Rand.) settings, and average denoising steps (Steps) for AnyStep-LingBotVA. The final row reports the aggregate results. Bold indicates the highest SR within each row and setting, including ties.}
\label{tab:lingbotva_all_tasks}
\vspace{\baselineskip}
\setlength{\tabcolsep}{4pt}
\renewcommand{\arraystretch}{0.94}
\begin{tabular}{@{}lcccccc@{}}
\toprule
Task & \multicolumn{2}{c}{LingBotVA~\citep{li2026causal}} & \multicolumn{4}{c}{AnyStep-LingBotVA} \\
\cmidrule(lr){2-3}\cmidrule(lr){4-7}
& Clean & Rand. & \multicolumn{2}{c}{Clean} & \multicolumn{2}{c}{Rand.} \\
\cmidrule(lr){4-5}\cmidrule(lr){6-7}
& SR $\uparrow$ & SR $\uparrow$ & SR $\uparrow$ & Steps $\downarrow$ & SR $\uparrow$ & Steps $\downarrow$ \\
\midrule
Adjust Bottle & 90 & 94 & \textbf{100} & 3.36 & \textbf{96} & 4.00 \\
Beat Block Hammer & 96 & \textbf{98} & \textbf{97} & 2.93 & \textbf{98} & 4.15 \\
Blocks Ranking RGB & \textbf{99} & \textbf{98} & 94 & 3.28 & 93 & 3.86 \\
Blocks Ranking Size & 94 & \textbf{96} & \textbf{97} & 3.27 & 81 & 3.81 \\
Click Alarmclock & 99 & \textbf{100} & \textbf{100} & 3.72 & \textbf{100} & 4.65 \\
Click Bell & \textbf{100} & \textbf{100} & \textbf{100} & 3.95 & \textbf{100} & 4.60 \\
Dump Bin Bigbin & 89 & 96 & \textbf{97} & 3.71 & \textbf{98} & 3.69 \\
Grab Roller & \textbf{100} & \textbf{100} & \textbf{100} & 3.34 & \textbf{100} & 4.22 \\
Handover Block & 99 & 78 & \textbf{100} & 3.46 & \textbf{95} & 3.43 \\
Handover Mic & 94 & \textbf{96} & \textbf{97} & 3.57 & 95 & 3.57 \\
Hanging Mug & \textbf{40} & 28 & 29 & 3.16 & \textbf{33} & 3.79 \\
Lift Pot & \textbf{100} & 99 & \textbf{100} & 2.80 & \textbf{100} & 3.99 \\
Move Can Pot & 94 & \textbf{97} & \textbf{97} & 3.27 & \textbf{97} & 3.89 \\
Move Pillbottle Pad & 99 & \textbf{99} & \textbf{100} & 3.25 & 98 & 3.88 \\
Move Playingcard Away & \textbf{100} & 99 & \textbf{100} & 3.61 & \textbf{100} & 4.25 \\
Move Stapler Pad & \textbf{91} & \textbf{79} & 67 & 3.57 & 71 & 3.56 \\
Open Laptop & 92 & \textbf{94} & \textbf{97} & 3.63 & \textbf{94} & 3.67 \\
Open Microwave & \textbf{82} & \textbf{86} & 69 & 3.14 & 85 & 3.82 \\
Pick Diverse Bottles & 89 & 82 & \textbf{92} & 3.87 & \textbf{89} & 3.66 \\
Pick Dual Bottles & \textbf{100} & \textbf{99} & \textbf{100} & 3.41 & 95 & 3.62 \\
Place A2B Left & 97 & \textbf{93} & \textbf{100} & 3.80 & 90 & 3.84 \\
Place A2B Right & \textbf{97} & \textbf{95} & \textbf{97} & 3.75 & 94 & 3.90 \\
Place Bread Basket & \textbf{97} & \textbf{95} & 94 & 3.66 & 90 & 3.54 \\
Place Bread Skillet & 95 & 90 & \textbf{97} & 3.87 & \textbf{95} & 3.75 \\
Place Burger Fries & \textbf{97} & 95 & \textbf{97} & 3.67 & \textbf{100} & 3.59 \\
Place Can Basket & 81 & 84 & \textbf{89} & 3.47 & \textbf{85} & 3.96 \\
Place Cans Plasticbox & \textbf{100} & \textbf{99} & \textbf{100} & 3.59 & 98 & 3.57 \\
Place Container Plate & \textbf{99} & \textbf{97} & 97 & 3.88 & 94 & 3.85 \\
Place Dual Shoes & \textbf{94} & \textbf{89} & 89 & 3.53 & 83 & 3.45 \\
Place Empty Cup & \textbf{100} & \textbf{100} & \textbf{100} & 3.68 & \textbf{100} & 3.85 \\
Place Fan & \textbf{99} & \textbf{93} & 97 & 3.86 & \textbf{93} & 3.70 \\
Place Mouse Pad & \textbf{93} & \textbf{96} & 92 & 3.73 & 94 & 3.84 \\
Place Object Basket & \textbf{91} & 88 & 88 & 3.45 & \textbf{93} & 3.45 \\
Place Object Scale & 96 & \textbf{96} & \textbf{97} & 3.27 & 92 & 3.85 \\
Place Object Stand & 99 & \textbf{96} & \textbf{100} & 3.88 & 94 & 3.88 \\
Place Phone Stand & 97 & \textbf{97} & \textbf{98} & 3.94 & 94 & 3.92 \\
Place Shoe & \textbf{98} & \textbf{98} & 96 & 3.87 & \textbf{98} & 3.88 \\
Press Stapler & \textbf{85} & 82 & 82 & 3.67 & \textbf{96} & 3.97 \\
Put Bottles Dustbin & \textbf{87} & \textbf{91} & 73 & 3.15 & 79 & 3.74 \\
Put Object Cabinet & 85 & 87 & \textbf{92} & 3.42 & \textbf{91} & 3.46 \\
Rotate Qrcode & 96 & 91 & \textbf{98} & 3.81 & \textbf{93} & 3.73 \\
Scan Object & \textbf{96} & \textbf{91} & 90 & 3.60 & \textbf{91} & 3.71 \\
Shake Bottle & \textbf{100} & 97 & \textbf{100} & 3.67 & \textbf{100} & 4.26 \\
Shake Bottle Horizontally & \textbf{100} & \textbf{99} & \textbf{100} & 3.60 & 98 & 4.11 \\
Stack Blocks Three & 99 & 98 & \textbf{100} & 3.31 & \textbf{100} & 3.91 \\
Stack Blocks Two & \textbf{100} & 98 & \textbf{100} & 3.48 & \textbf{100} & 3.49 \\
Stack Bowls Three & \textbf{86} & \textbf{83} & 84 & 3.26 & 79 & 3.79 \\
Stack Bowls Two & 94 & \textbf{98} & \textbf{100} & 3.48 & 96 & 3.44 \\
Stamp Seal & \textbf{96} & \textbf{97} & \textbf{96} & 3.85 & 96 & 3.93 \\
Turn Switch & 44 & 45 & \textbf{61} & 3.50 & \textbf{61} & 3.53 \\
\midrule
\textbf{Average} & \textbf{92.93} & 91.55 & 92.74 & 3.54 & \textbf{91.7} & 3.82 \\
\bottomrule
\end{tabular}
\end{table}

\end{document}

%% file: math_commands.tex
\usepackage{amsmath,amsfonts,bm}

\def\eqref#1{equation~\ref{#1}}

\def\1{\bm{1}}

\def\vtheta{{\bm{\theta}}}
\def\va{{\bm{a}}}

\def\vc{{\bm{c}}}

\def\vl{{\bm{l}}}

\def\vo{{\bm{o}}}

\def\vu{{\bm{u}}}
\def\vv{{\bm{v}}}

\def\vx{{\bm{x}}}

\def\vz{{\bm{z}}}

\DeclareMathAlphabet{\mathsfit}{\encodingdefault}{\sfdefault}{m}{sl}
\SetMathAlphabet{\mathsfit}{bold}{\encodingdefault}{\sfdefault}{bx}{n}

